%% file: main.tex
\documentclass[conference]{IEEEtran}
\usepackage{times}

\usepackage[numbers]{natbib}
\usepackage{multicol}
\usepackage[bookmarks=true]{hyperref}

\usepackage{booktabs}
\usepackage{float}
\usepackage{graphicx}
\usepackage{amsfonts}
\usepackage{bbding}
\usepackage{multirow}
\usepackage{makecell}
\usepackage{caption}
\usepackage{amsmath}
\usepackage{xcolor}

\newcommand{\specialcell}[2][c]{%
  \begin{tabular}[#1]{@{}c@{}}#2\end{tabular}}

\begin{document}

\title{Any-point Trajectory Modeling for Policy Learning}

\author{Chuan Wen $^{*1,2,5}$ \quad Xingyu Lin$^{*1}$ \quad John So$^{*3}$ \\
Kai Chen$^{6}$ \quad Qi Dou$^{6}$ \quad Yang Gao$^{2,4,5}$ \quad Pieter Abbeel$^{1}$ \\
$^{1}$UC Berkeley \quad $^{2}$IIIS, Tsinghua University \quad $^{3}$Stanford University \\
$^{4}$Shanghai AI Laboratory \quad $^{5}$Shanghai Qi Zhi Institute \quad $^{6}$ CUHK \\
}



%

\twocolumn[{%
\renewcommand\twocolumn[1][]{#1}%
\maketitle
\begin{center}
    \centering
    \captionsetup{type=figure}
    \includegraphics[width=0.92\textwidth]{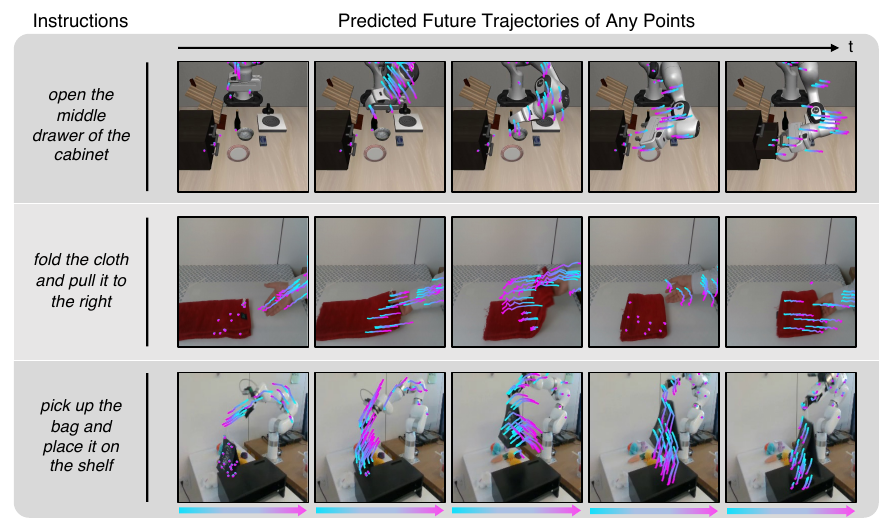}
    \captionof{figure}{Given a task instruction and the initial positions of any set of points in an image frame, our \textbf{A}ny-point \textbf{T}rajectory \textbf{M}odel (ATM) can predict the future trajectories of these points conditioned on the task. After training the model on an action-free video dataset, the predicted trajectories serve as effective guidance for learning visuomotor policies for a set of language-conditioned manipulation tasks.}
\end{center}%
}]

\IEEEpeerreviewmaketitle

\makeatletter\def\Hy@Warning#1{}\makeatother\let\thefootnote\relax\footnotetext{*First three authors contributed equally: Chuan Wen led the implementation and experiments. 
Xingyu Lin came up with the idea, supervised the technical development, and contributed to model debugging. 
John So implemented the Robot-to-robot transfer experiments and UniPi baselines.}

\input{sec/0_abstract}    
\input{sec/1_intro}
\input{sec/2_related_work}
\input{sec/3_method}

\input{sec/4_experiment}
\input{sec/5_conclusion}

\bibliographystyle{plainnat}
\bibliography{references}

\input{sec/6_supp}

\end{document}

%% file: sec/0_abstract.tex
\begin{abstract}
Learning from demonstration is a powerful method for teaching robots new skills, and having more demonstration data often improves policy learning. However, the high cost of collecting demonstration data is a significant bottleneck. Videos, as a rich data source, contain knowledge of behaviors, physics, and semantics, but extracting control-specific information from them is challenging due to the lack of action labels. In this work, we introduce a novel framework, \textbf{A}ny-point \textbf{T}rajectory \textbf{M}odeling (ATM), that utilizes video demonstrations by pre-training a trajectory model to predict future trajectories of arbitrary points within a video frame. Once trained, these trajectories provide detailed control guidance, enabling the learning of robust visuomotor policies with minimal action-labeled data. Across over \textbf{130} language-conditioned tasks we evaluated in both simulation and the real world, ATM outperforms strong video pre-training baselines by 80$\%$ on average. Furthermore, we show effective transfer learning of manipulation skills from human videos and videos from a different robot morphology. Visualizations and code are available at: \url{https://xingyu-lin.github.io/atm}.
\end{abstract}

%% file: sec/1_intro.tex
\section{Introduction} \label{sec:intro}
Computer vision and natural language understanding have made significant advances in recent years~\cite{kirillov2023segment,brown2020language}, where the availability of large datasets plays a critical role. Similarly, in robotics, scaling up human demonstration data has been key for learning new skills~\cite{brohan2022rt,padalkar2023open,fang2023rh20t}, with a clear trend of performance improvement with larger datasets~\cite{mandlekar2021matters,brohan2022rt}. However, human demonstrations, typically action-labeled trajectories collected via teleoperation devices~\cite{zhang2018deep,wu2023gello}, are time-consuming and labor-intensive to collect. For instance, collecting $130K$ trajectories in~\cite{brohan2022rt} took 17 months, making data collection a major bottleneck in robot learning.

Videos contain knowledge about behaviors, physics, and semantics, presenting an alternative data source. However, the lack of action labels makes utilization of video data in policy learning difficult. Previous works have addressed this by using self-supervised objectives for video pre-training to learn a feature representation of the observation for policy learning~\cite{sermanet2018tcn,nair2023r3m,seo2022reinforcement}. However, a feature representation only describes the state at the current time step, largely neglecting the transition dynamics that predicts future states. To explicitly model the transition dynamics, prior works have developed video prediction models that predict future image frames from current ones to guide policy learning~\cite{du2023unipi,yang2023learning, Escontrela23arXiv_VIPER}. However, learning a video prediction model for control introduces two challenges. Firstly, the task of video prediction avoids any abstraction by modeling changes to every pixel, coupling the physical motion with visual appearances such as texture, and lighting. This coupling makes modeling difficult, often resulting in hallucinations and unrealistic future predictions~\cite{du2023unipi}. Secondly, these models are computationally demanding in both training and inferencing. With limited computational resources, performance significantly declines. Moreover, the high inference cost compels these models to adopt open-loop execution~\cite{du2023unipi,black2023zero}, which tends to result in less robust policies.

In this paper, we propose a novel and structured representation to bridge video pre-training and policy learning. We first represent each state as a set of points in a video frame. To model the temporal structure in videos, we learn an Any-point Trajectory Model~(ATM) that takes the positions of the points in the current frame as input and outputs their future trajectories. We predict these trajectories in the camera coordinate frame to minimize the assumptions on calibrated cameras. These 2D point trajectories correspond to trajectories of particles in the 3D space and are a universal representation of the motions that can transfer to different domains and tasks. Contrasting with the video prediction approach of tracking changes in pixel intensity, the particle-based trajectory modeling offers a faithful abstraction of the physical dynamics, naturally incorporating inductive biases like object permanence and continuous motion. We first pre-train the trajectory model on actionless video datasets. After pre-training, the predicted trajectories can function as subgoals to guide the control policy. We then train trajectory-guided policies using only a minimal amount of action-labeled demonstration data. For training the ATMs, we generate self-supervised training data by leveraging recent advancements in vision models for accurate point tracking~\cite{karaev2023cotracker}. Across over \textbf{130} language-conditioned tasks we evaluated in both simulation and the real world, ATM significantly surpasses various strong baselines in video pre-training, achieving an average success rate of $63\%$ compared to the highest success rate of $37\%$ by previous methods, marking an improvement of over $80\%$. Additionally, we demonstrate that our method facilitates effective transfer learning from human videos and videos of a robot with a different morphology. We summarize our main contributions below:
\begin{enumerate}
  \item We propose an Any-point Trajectory Model, a simple and novel framework that bridges video pre-training to policy learning, leveraging the structured representation of particle trajectories.
  \item Through extensive experiments on simulated benchmarks and in the real world, we demonstrate that our method can effectively utilize video data in pre-training and significantly outperform various video pre-training baselines in an imitation learning setting.
  \item We demonstrate effective learning from cross-embodiment human and robot videos.
\end{enumerate}

%% file: sec/2_related_work.tex
\section{Related Work}
\noindent\textbf{State representation for control.}
In learning end-to-end visuomotor policies, the policy is typically parameterized as a neural network that takes image observation as the input representation~\cite{reed2022generalist,brohan2022rt,walke2023bridgedata}. Due to the lack of inductive bias, these approaches require training on a large number of demonstration trajectories, which is expensive to collect. On the other hand, different structured representations are proposed to improve the data efficiency, such as key points~\cite{qin2020keto,manuelli2020keypoints}, mesh~\cite{lin2021VCD,huang2022medor}, or neural 3D representation~\cite{li20213d,rashid2023language}. However, prior structures often limit the policy to specific tasks. In contrast, we propose to utilize future trajectories of arbitrary points in the image as additional input to the policy, making minimal assumptions about the task and environment. We demonstrate its wide application to a set of over 130 language-conditioned manipulation tasks. In navigation and locomotion, it is common to construct policies that are guided by the future trajectories of the robot~\cite{aguiar2007trajectory,peng2018deepmimic}. In manipulation, some works have explored flow-based guidance~\cite{goyal2022ifor,seita2023toolflownet,gu2023rt}. However, prior works only track task-specific points, such as the end-effector of the robot, or the human hand. Instead, our approach works with arbitrary points, including points on the objects, thus providing richer information in the more general settings. Finally, \citet{vecerik2023robotap} proposes to utilize any-point tracking for few-shot policy learning. This approach does not learn trajectory models from data, but instead mainly uses the tracker to perform visual servoing during test time. This design choice requires more instrumentation, such as separating the task into multiple stages, predicting the goal locations of the points for each stage, and running the tracker during inference time. In contrast, we present a much simpler framework, enabling application in more diverse settings.

\noindent\textbf{Video pre-training for control.} Videos contain rich information about behaviors and dynamics, which can help policy learning. However, video pre-training remains challenging due to the lack of action labels. One line of works first learns an inverse dynamics model that predicts the action from two adjacent frames and then labels the videos with pseudo actions~\cite{schmeckpeper2019learning, baker2022vpt, torabi2018behavioral,schmidt2024learning}. However, the inverse dynamics model is often trained on a limited action-labeled dataset and does not generalize well, especially for continuous actions. Prior works have also explored pre-train a feature representation using various self-supervised objectives~\cite{sermanet2017tcn,nair2023r3m,ma2023vip}, but the representation alone does not retain the temporal information in videos. More recently, learning video prediction as pre-training has shown promising results~\cite{du2023unipi,Ko2023Learning,bharadhwaj2023zero,yang2023learning,Escontrela23arXiv_VIPER}. During policy learning, a video prediction model is used to generate future subgoals and then a goal-conditioned policy can be learned to reach the sub-goal. However, video prediction models often result in hallucinations and unrealistic physical motions. Furthermore, video models require extensive computation, which is an issue, especially during inference time. In contrast, our method models the trajectories of the points, naturally incorporating inductive bias like object permanence while requiring much less computation. This enables our trajectory models to be run closed-loop during policy execution. Furthermore, the trajectories provide dense guidance to the policy as a motion prior.

\noindent\textbf{Learning from Human Videos.} Of particular interest, numerous prior works have proposed learning from the rich source of human videos~\cite{xiong2021learning,shao2021concept2robot, mendonca2023structured,bahl418affordances,shaw2023videodex,wang2023mimicplay}. However, these works often explicitly extract the hand pose or contact regions from the human videos, thereby losing information about the dynamics of the remaining objects. In contrast, our method models the trajectories of arbitrary points and can learn from both human videos and videos of a different robot.

%% file: sec/3_method.tex
\begin{figure*}
    \centering
  \includegraphics[width=\linewidth]{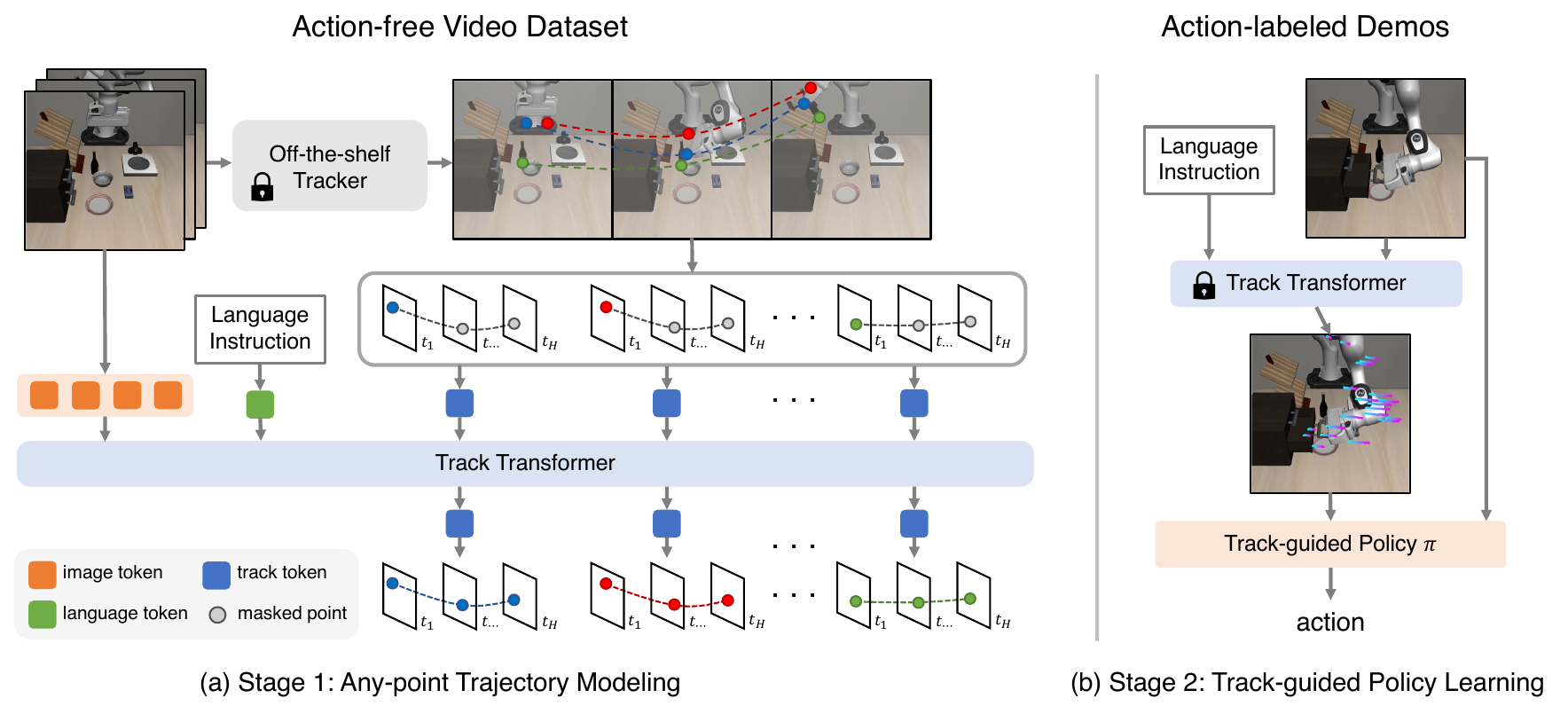}
  \caption{Overview of our framework. (a) In the first stage, given an action-free video dataset, we first sample 2D points on one video frame and track their trajectories throughout the video using a pre-trained tracker. We then train a track transformer to predict future point trajectories given the current image observation, the language instruction, and the initial positions of the points. For the transformer input, we replace the future point positions with masked values. (2) In the second stage, we learn a track-guided policy to predict the control actions. Guidance from the predicted track enables us to learn robust policies from only a few action-labeled demonstrations. }
  \label{fig:system}
\end{figure*}

\section{Preliminary} \label{sec:prelim}
In this paper, we aim to learn robust control policies from a small set of action-labeled demonstration trajectories. Our central goal is to leverage the more scalable, unlabeled videos as a data source for pre-training. 

\noindent \textbf{Imitation from demos and videos.} To begin with, we denote the action-free video dataset as $\mathcal{T}_{o}=\{(\tau_o^{(i)}, \ell^{(i)})\}_{i=1}^{N_{o}},$ where $\ell^{(i)}$ is the language instruction for the $i^{th}$ episode and $\tau_o^{(i)} = \{o_t^{(i)}\}_{t=1}^T$ denotes the observation-only trajectory consisting of camera images. Similarly, we denote the demonstration dataset as $\mathcal{T}_{a}=\{(\tau_a^{(i)}, \ell^{(i)})\}_{i=1}^{N_{a}},$ where $\tau_a^{(i)} = \{o_t^{(i)}, a_t^{(i)}\}_{t=1}^T$ is the action-labeled trajectory. During imitation learning, our goal is to learn a policy $\pi_{\theta}$, parameterized by $\theta$ to mimic the expert behavior by the following behavioral cloning objective:
\begin{equation}
    \pi_{\theta}^{*} = \arg\min_{\theta} E_{(o_t,a_t, \ell)\sim \mathcal{T}_{a}}\Big[\mathcal{L}\Big(\pi_{\theta}(o_t, \ell), a_t\Big)\Big],
    \label{eq:bc}
\end{equation}
where $\mathcal{L}$ is the loss function, which could be Mean Squared Error (MSE) or cross-entropy loss.

\noindent \textbf{Tracking Any Point (TAP).} The recent advancements in video tracking~\cite{harley2022particle, doersch2023tapvid, doersch2023tapir, wang2023tracking, zheng2023pointodyssey} enable us to track the trajectory of each point in video frames without external supervision. In this paper, we utilize the off-the-shelf tracker proposed in~\cite{karaev2023cotracker}. Formally, given a sequence of images from a video $o_1, ..., o_T$, any one of the time steps $\bar{t} \in [1, T]$, and a set of points in that frame $\{p_{\bar{t}, k}\}_{k=1}^{K}$, where $p_{\bar{t}, k} = (x, y)$ is the point coordinate in the camera frame, the task of tracking is to predict the 2D camera-frame coordinates of the corresponding points in every frame $p_{t, k}$ where $t =1 \dots T$. In this paper, we use the terms \textit{trajectory} and \textit{track} interchangeably to refer to a sequence of 2D coordinates of any point, denoted as $(p_1, \dots, p_T)$. We only model the 2D trajectories in the camera frame so that we do not have to make additional assumptions about multi-view cameras, or the availability of depth, allowing future scaling to more diverse video datasets. The tracker additionally predicts a binary visibility value $v_{t, k}$ denoting whether the point is occluded at step $t$.

\section{Method}
Videos contain a great deal of prior information about the world, capturing physical dynamics, human behaviors, and semantics that are invaluable for policy learning. Beyond just learning representations from videos~\cite{sermanet2017tcn,nair2023r3m,ma2023vip}, we aim to learn a model from videos to predict future states for guiding a control policy. In this way, we can essentially decompose the visuomotor policy learning challenge into two parts. The first part is learning what to do next by generating future states as concrete \textit{sub-goals}, which is learned purely from videos. The second part is learning to predict control actions to follow the sub-goals, which require much less data to train compared to learning policies end-to-end. With sufficient video pre-training, we will be able to learn generalizable policies even from limited action-labeled trajectories. Prior works~\cite{du2023unipi,Ko2023Learning,black2023zero} have predominantly relied on pixel-level future frame prediction as video pre-training. While video prediction is resource-intensive during both training and inference stages, its focus on reconstructing pixel-level details, which are often extraneous to policy learning, can adversely affect the efficiency of subsequent policy learning.

We propose \textbf{A}ny-point \textbf{T}rajectory \textbf{M}odeling~(\textbf{ATM}). As illustrated in Figure~\ref{fig:system}, 
ATM is a two-stage framework: first learn to predict future point trajectories in a video frame as the pre-training with large-scale action-free videos, then use the predicted trajectories to guide policy learning with few-shot action-labeled demos. 
Our proposed method will be comprehensively detailed in this section: In Sec.~\ref{sec:track-model}, we first describe how to learn a point trajectory prediction model from an action-free video dataset $\mathcal{T}_{o}$. Then in Sec.~\ref{sec:track-policy}, we outline how we utilize the pre-trained track prediction model to learn a track-guided policy from a limited action-labeled trajectory datasets $\mathcal{T}_a$.

\subsection{Trajectory Modeling from Video Datasets} \label{sec:track-model}
Our goal is to pre-train a model from videos that forecasts the future point trajectories in a frame. More formally, given an image observation $o_t$ at timestep $t$, any set of 2D query points on the image frame $\mathbf{p}_t = \{p_{t, k}\}_{k=1}^K$, and a language instruction $\ell$, we learn a model $\mathbf{p}_{t:t+H} = \tau_\theta(o_t, \mathbf{p}_t, \ell)$ that predicts the coordinates of the query points in the future $H$ steps in the camera frame, where  $\mathbf{p}_{t:t+H} \in \mathbb{R}^{H \times K \times 2}$. To model the tracks, we propose a \textbf{track transformer} and illustrate the architecture in Figure~\ref{fig:system} (a).

\paragraph{Self-supervised Track Annotation.} \label{sec:TAP-annotation}
Initially, we generate point trajectories from action-free videos for trajectory modeling pre-training.
As described in Sec.~\ref{sec:prelim}, we employ a vision tracker to pre-process videos and generate a tracking dataset. For each video, we randomly sample a time step $\bar{t}$ and then randomly sample points on this frame and generate their tracks by running the tracker. However, for a static camera, most of the points that are sampled randomly will be in the background, thus providing little information when training our track transformer. To address this, we adopt a heuristic solution to filter out these static points: we first sample a grid of $n~\times~n$ on frame $\bar{t}$ and track the grid of points across the whole of the video to obtain an initial set of tracks $\tau \in \mathbb{R}^{n^2 \times T \times 2}$. Subsequently, we filter points that have not moved during the video by thresholding the variance of the point positions over time. In the final step, we resample points around the filtered locations and finally generate their positions using the tracker.

\paragraph{Multimodal Track Modeling.}
We formalize the future forecasting problem as a multi-modal masked prediction problem: we aim to predict the future positions of each point, conditioned on its current position, the current image observation, and a language instruction of the task. We first encode different modalities into a shared embedding space, each represented by a few tokens. For the tracks, we mask out the future positions of all points before encoding and then separately encode each point into one token. For the language instruction, we use a pre-trained BERT~\cite{devlin2019-bert} encoder. For the images, we split them into image patches and randomly masked out $50\%$ of the patches. We then pass all tokens through a large transformer model. Finally, we decode the track tokens into future trajectories of the corresponding points. Additionally, we reconstruct the image patches from the corresponding tokens following~\citet{he2022masked} as an auxiliary task, which we find useful for more complex tasks. Through this pre-training process, our track transformer learns the motion prior of particles within the video frames.

\subsection{Track-guided Policy Learning}\label{sec:track-policy}
After training a track transformer to predict future tracks based on observations, we can then learn policies guided by these predicted trajectories.

\paragraph{Arbitrary Points Tracking.} During track transformer pre-training, we can filter tracks without large movements. However, using this heuristic requires knowing the future positions of each point, which can be expensive to compute during policy inference. Instead, we find it sufficient to simply use a fixed set of 32 points on a grid for the policy. This sampling method avoids the potential complexities of learning key points or finding points to track~\cite{vecerik2023robotap} and works well in practice. ATM is permutation invariant to the input set of points, and we also find ATM to be robust to the distribution of the points, allowing us to use a different point sampling scheme from training for policy learning.

\begin{figure}[ht]
  \centering
  \includegraphics[width=.75\linewidth]{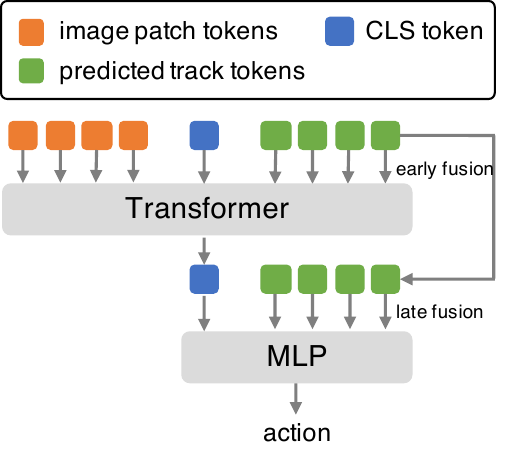}
  \caption{A visual illustration of the architecture of the track-guided policy. Given the current observation and the predicted tracks from the frozen pre-trained track transformer, we train a track-guided policy from a limited demonstration dataset.}
  \label{fig:track_policy}  
\end{figure}

\paragraph{Track-guided Policy Learning.} A track-guided policy $\pi(a_t | o_t, \textbf{p}_{t:t+H})$ takes input the current observation $o_t$ and the predicted tracks $\textbf{p}_{t:t+H}$ and predict the actions. A simplified illustration of our policy architecture is shown in Figure~\ref{fig:track_policy}. Our transformer policy architecture follows the architecture in prior works~\cite{liu2023libero, kim2021vilt}. Although the predicted tracks alone already provide rich information to predict the actions, we still incorporate contextual image observations into our policy so that no information is lost, as suggested in prior works~\cite{lin2023spawnnet}. We incorporate the track tokens both before and after fusion with the image tokens (early fusion and late fusion) to ensure that the guiding information from tracks can be effectively integrated. Surprisingly, as the tracks already provide the fine-grained subgoals, we find that the policy no longer needs language instruction at this stage as task specification. Essentially, the provided tracks have reduced the difficult policy learning problems into a much easier sub-goal following problems, reducing the policy into an inverse dynamics model. Our track-guided policy is trained with MSE loss. 
Note that, the weights of our policy model are randomly initialized rather than copied from the pretrained Track Transformer like other video-pretraining methods~\cite{nair2023r3m,ma2023vip}, because of the aim to separately study the task guidance capability of predicted trajectories.
A detailed architecture diagram and hyperparameters are available in the appendix.

%% file: sec/4_experiment.tex
\section{Experiments}\label{sec:experiment}
We perform experiments to answer the following questions:
\begin{itemize}
    \item How does ATM compare with state-of-the-art video pre-training and behaviour cloning baselines for learning from action-free videos?
    \item Can ATM be used for learning from video data that are out of the distribution of demonstration data?
\end{itemize}
Our experiments are split into three sections. In Sec.~\ref{sec:video_bc}, we compare ATM with video pre-training baselines on over 130 language-conditioned manipulation tasks in simulation and in the real world. In Sec.~\ref{sec:human_video}, we show that ATM enables effective learning from human videos. Finally, we present ablation results in Sec.~\ref{sec:ablation}.

\begin{figure*}[ht]
    \centering
    \includegraphics[width=\linewidth]{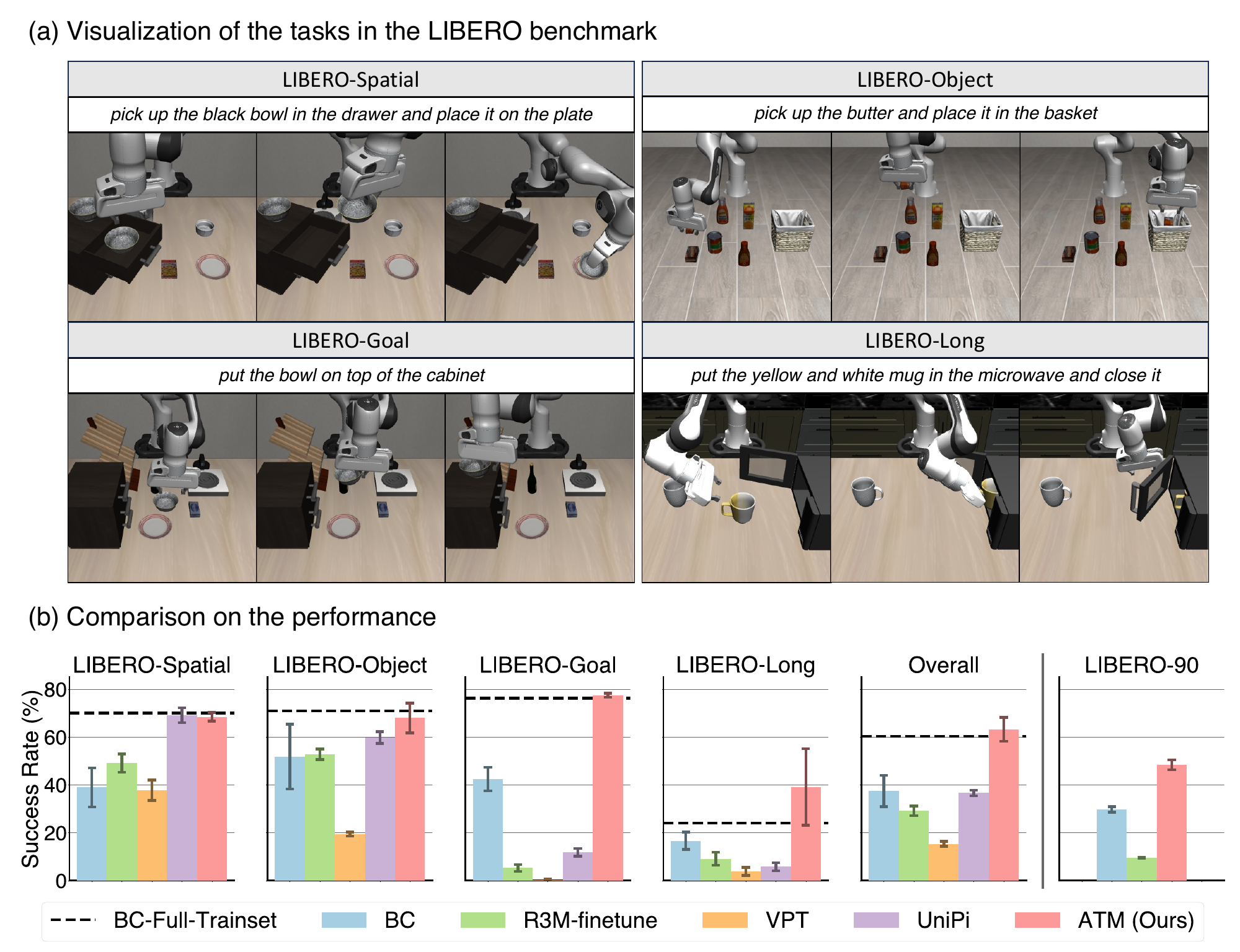}
    \caption{We compare with state-of-the-art video pre-training methods on language-conditioned manipulation tasks in the LIBERO benchmark~\cite{liu2023libero}. (a) Visualization of the LIBERO tasks separated into four suites, focusing on different aspects of the manipulation policies in spatial reasoning, object reasoning, task understanding, and performing long-horizon tasks. (b) Quantitative comparisons on different suites. We additionally compare baselines with fast computation on a task suite containing 90 tasks (i.e. LIBERO-90). ATM outperforms the baselines in all tasks and excels in LIBERO-Goal and LIBERO-Long.}
    \label{fig:sim_exp}
\end{figure*}

\begin{figure*}[ht]
    \centering
    \includegraphics[width=\linewidth]{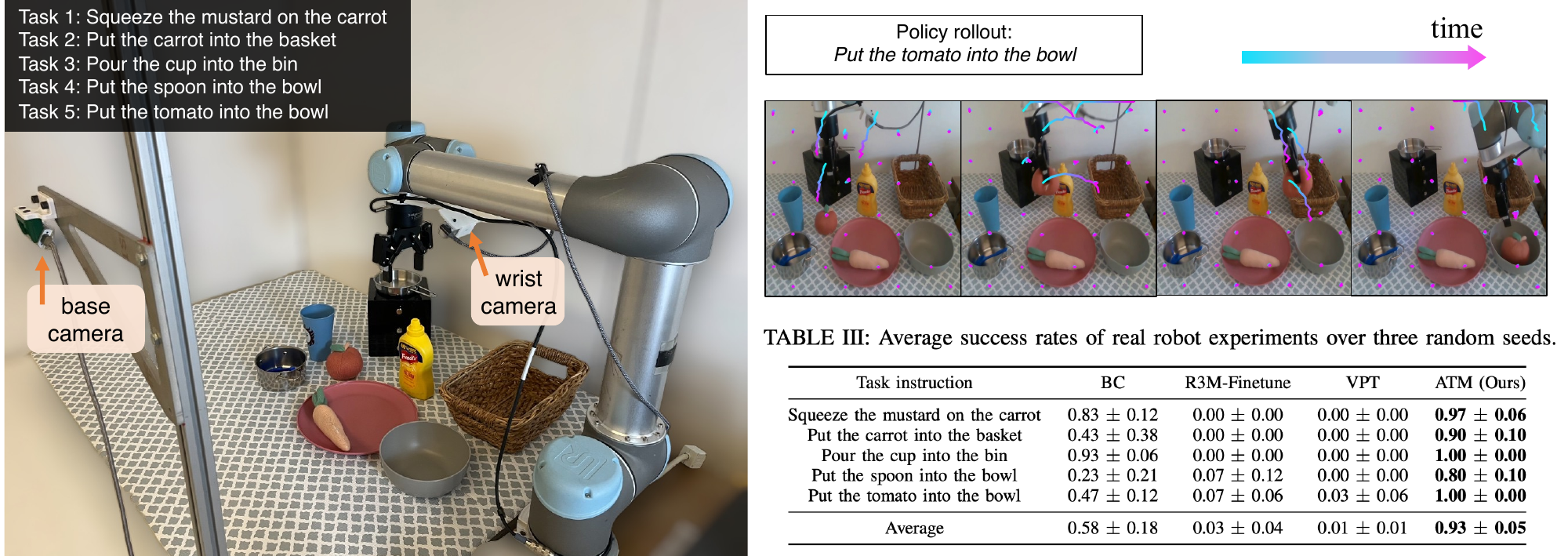}
    \caption{Real robot experiments on a dining table setup consisting of five tasks. The left figure shows our real-world setup and the tasks. The top right figure shows an example of the predicted particle trajectories and the policy execution, which closely follows the predicted trajectories. From the quantitative results, we can see that ATM shows significant improvements over state-of-the-art video pre-training baselines on average.}
    \label{fig:real_exp}
\end{figure*}

\subsection{Video Pre-training for Imitation Learning}~\label{sec:video_bc}
\noindent\textbf{Environments.} We compare with baselines on over one hundred language-conditioned manipulation tasks in the LIBERO benchmark~\cite{liu2023libero}. The benchmark is subdivided into five suites, LIBERO-Spatial, LIBERO-Object, LIBERO-Goal, LIBERO-Long, and LIBERO-90. Each suite has 10 tasks, except LIBERO-90 which contains 90 tasks. Each task comes with expert human demonstrations. LIBERO-Spatial contains tasks with the same objects but different layouts; LIBERO-Object has tasks with the same layouts but different objects; LIBERO-Goal has tasks with the same object categories and spatial layouts, but different goals; LIBERO-Long has tasks with diverse object categories and layouts, and long-horizon task goals; LIBERO-90 has extremely diverse object categories, layouts and task goals.

\textbf{Data.} We compare with baselines on each suite separately. All methods are trained on $10$ action-labeled demonstration trajectories and $50$ action-free video demonstration trajectories of the robot for each task, amounting to $500$ videos for each 10-task suite. The demonstration dataset contains RGB images from a third-person camera and a wrist camera, together with gripper and joint states as observations. As each task is specified by a language instruction, we use the pre-trained BERT network to obtain a task embedding~\cite{devlin2019-bert}. Image resolution is $128\times128$ and the action space is 7-dimension, representing the translation, rotation, and aperture of the end-effector. 

\textbf{Baselines.}
We compare with the following baselines:
\begin{enumerate}
    \item \textbf{BC} denotes the vanilla behavioral cloning which trains a policy exclusively using the limited action-labeled expert demonstrations, without using the video dataset. It uses a policy architecture identical to ATM except that the particle trajectories are masked to be zero and it instead takes language embedding as task specification. BC baseline can also be considered as an ablation for policy learning without action-free videos. The specific architecture and losses for training the BC policy can vary. We use BC to denote a transformer policy trained using MSE loss by default and compare with diffusion policy separately towards the end of this section.
    \item  \textbf{R3M-finetune}~\cite{nair2023r3m} uses a contrastive learning objective for learning representation that aligns video and language with a combination of time contrastive losses, $L_{1}$ regularization, and language consistency losses. We adopt the publicly released Ego4D pre-trained weights and fine-tune the weights on our in-domain video dataset $\mathcal{T}_o$, to initialize the behavioral cloning policy's visual encoder. During policy training, we also further fine-tune the R3M backbone with the behavioral cloning loss. While this method captures priors from action-free videos through representation learning, the visual representation lacks knowledge about the transition dynamics critical to decision-making.
    \item \textbf{VPT}~\cite{baker2022vpt} first trains an inverse dynamics model from the action-labeled trajectories $\mathcal{T}_a$ and then uses it to predict pseudo action labels for the video dataset. With these pseudo-labels, a policy is then trained with behavioral cloning. This method requires the inverse dynamics model to be robust to a wide distribution of input observation, which can be difficult to learn from the limited demonstrations. 
    \item \textbf{UniPi}~\cite{du2023unipi, Ko2023Learning,black2023zero} trains a text-conditioned video diffusion model to generate a temporally fine-grained video plan from an initial frame and a language instruction. During policy learning, UniPi trains an inverse dynamics model with action-labeled data. We base our implementation off of the UniPi implementation in~\citet{Ko2023Learning}. While both UniPi and ATM leverage a policy conditioned on future subgoals, a trajectory representation decouples motion from other pixel-based information and makes policy learning much easier. Please see the Appendix where we perform additional comprehensive comparisons of different variations of UniPi. 
\end{enumerate}

\textbf{Results.} We present the main results in Figure~\ref{fig:sim_exp}. We see that by bridging the video data and policy learning with the structured representation of point trajectories, ATM~(our method) significantly surpasses various strong baselines in video pre-training, achieving an average success rate of $63\%$ compared to the highest success rate of $37\%$ by previous methods, marking an improvement of over $80\%$. The comparison of BC with ATM shows that learning from additional videos provides useful information for policy learning. VPT performs poorly as we empirically observed that the pseudo-action labels predicted by VPT generally show large errors on the video dataset. UniPi fails on more complex as video prediction models are not physically grounded and often generate future frames that are not physically feasible, such as cases where robots disappear from the image. 
ATM shows surprisingly superior performance on long-horizon tasks and tasks that require an understanding of goals. We attribute this performance gain to ATM's formulation, which utilizes explicit future particle trajectories as subgoals. At each time step, ATM predicts a future trajectory goal based on the current observation. In comparison to the static natural language instructions used by other methods, which do not change throughout the episode, ATM's closed-loop future trajectory provides clear guidance as to what the policy needs to do next.
Please see our video for failure cases of a video prediction model.

\begin{figure}[ht]
    \centering
    \includegraphics[width=\linewidth]{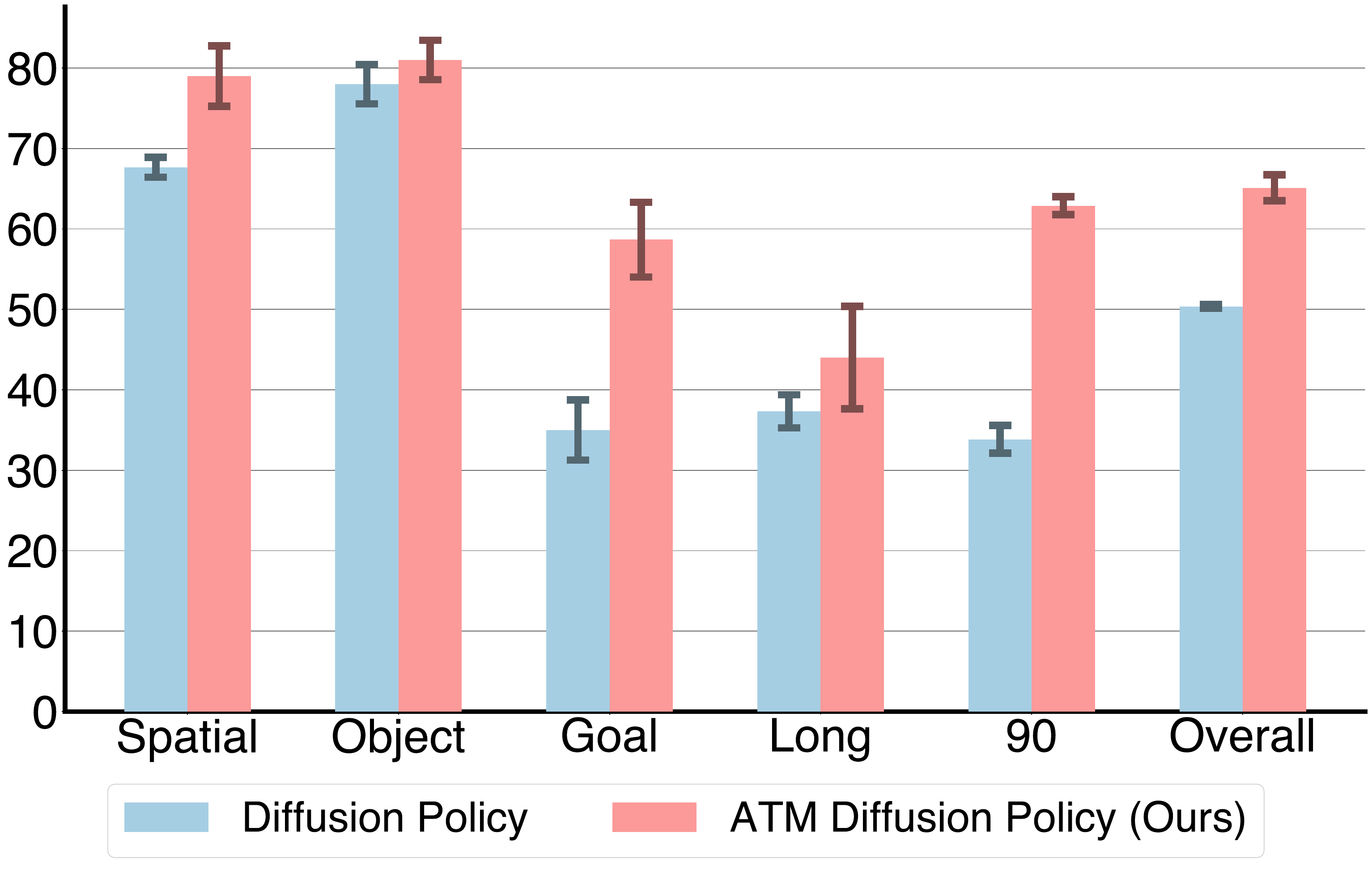}
    \caption{We implement ATM Diffusion Policy by adding the predicted future trajectories as additional conditioning and show consistent improvement over the base diffusion policies across the benchmark suites.}
    \label{fig:diff-policy-results}
\end{figure}

\noindent\textbf{Universality of ATM on Different Policy Classes.} The benefit of the predicted trajectories can also be shown for state-of-the-art policy class such as diffusion policy~\cite{Chi2023diffusion}, as shown in Figure~\ref{fig:diff-policy-results}. Here, ATM Diffusion Policy adds the predicted future tracks as additional condition to the policy class, while keeping everything else the same.  We see that ATM consistently improves the performance of diffusion policy across all the suites, achieving the highest scores on the LIBERO benchmark. More details can be found in Appendix.

\textbf{Real World Setup.} We conduct a language-conditioned manipulation experiment in the real world to further strengthen our claim. As illustrated in Figure~\ref{fig:real_exp}, we learn policies for a 6-DOF UR5 robot arm using human expert demonstrations collected with the GELLO teleoperation system~\cite{wu2023gello}. The action space is joint position control and gripper state. The observations include two RGB images from two RealSense cameras. To compensate for the partial observation, we stack the most recent two frames as the input of agents. We do not feed proprioception states into agents because it is reported that imitation policies tend to overfit them and ignore the visual inputs, leading to worse online rollout performance~\cite{robomimic2021,lin2023spawnnet}. We collect a total of $50$ action-labeled trajectories and $250$ action-free demonstration videos (by simply removing the actions from the collected action-labeled demonstrations). From the quantitative results in Figure~\ref{fig:real_exp}, we see consistent trends that ATM outperforms BC and other video pre-training baselines by a significant margin.

\begin{figure*}[ht]
    \centering
    \includegraphics[width=\linewidth]{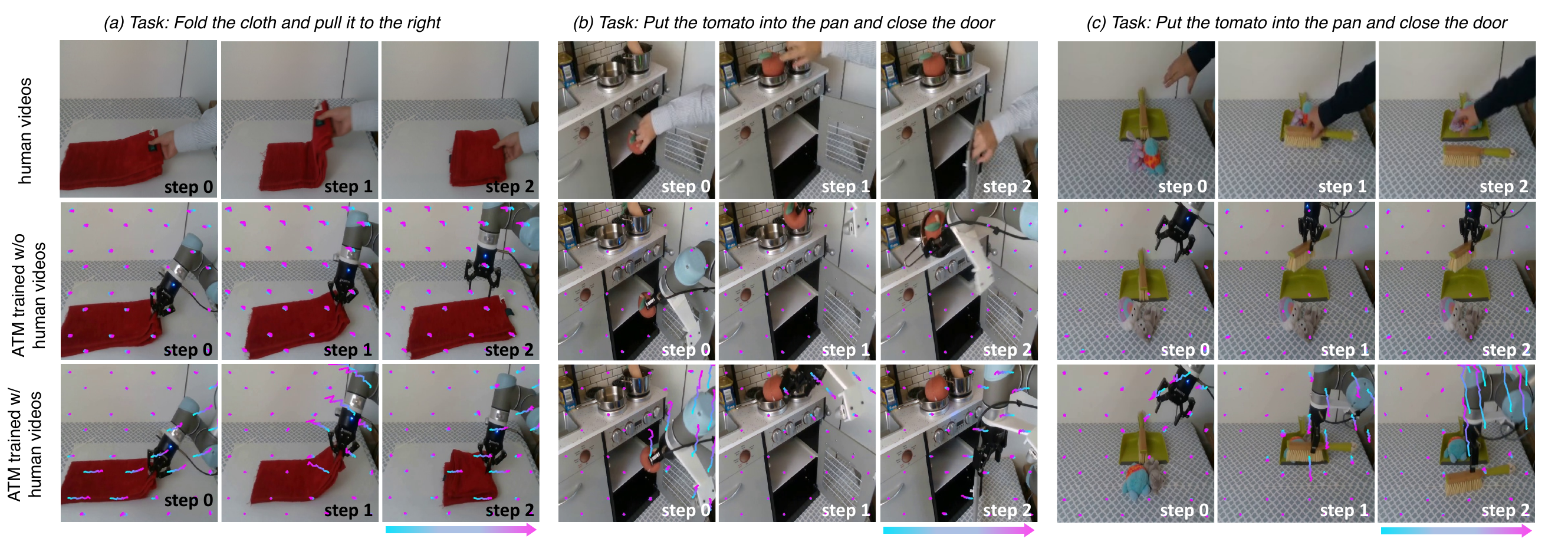}
    \caption{Learning robotic skills from human videos for three tasks. We collect 100 videos of a human performing the tasks directly and 10 teleoperation demonstration trajectories. Each row from the top to the bottom shows three snapshots from the human videos, ATM trained without the human videos, and ATM trained with the human videos. By comparison, we can see that human videos are critical in learning accurate trajectory prediction and enable the policy to successfully perform the task.}
    \label{fig:human_video}
\end{figure*}

\begin{figure*}[ht]
    \centering
    \includegraphics[width=\linewidth]{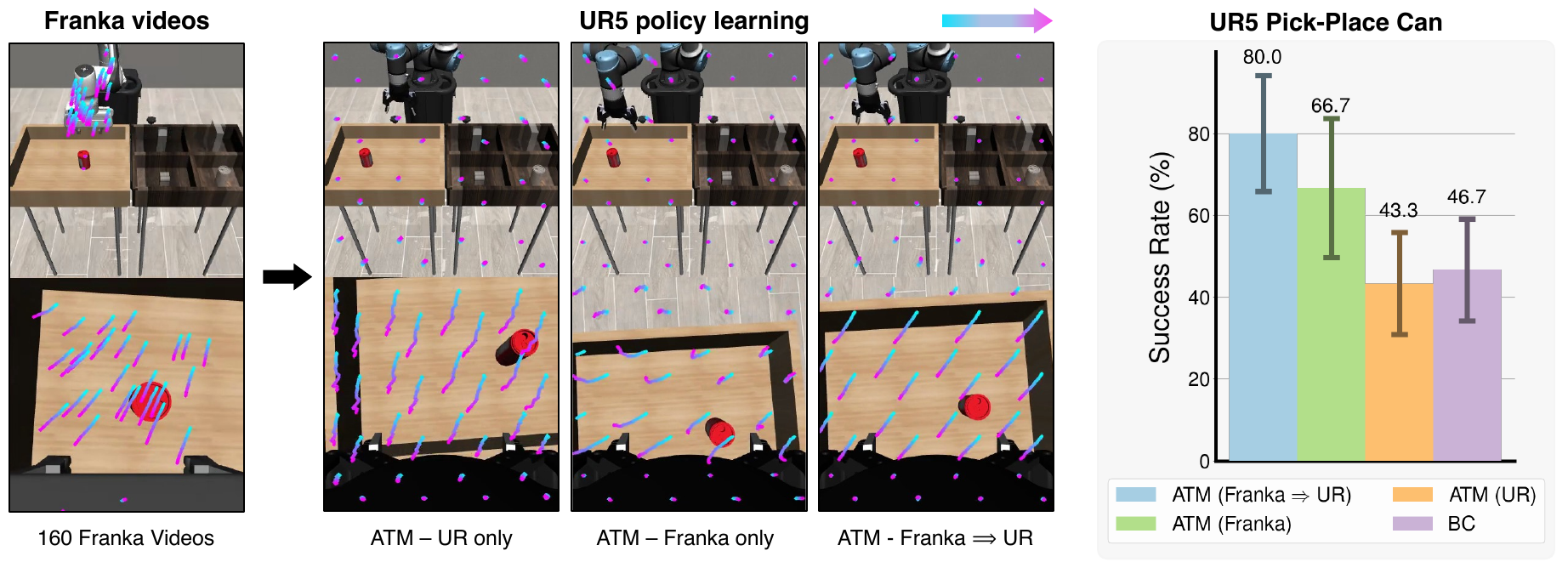}
    \caption{Cross-morphology skill transfer for a pick-and-place task. Here, we collect 160 action-free videos of a Franka arm and 10 action-labeled demonstrations from a UR arm, with the final goal of learning a UR policy. We compare a vanilla BC baseline with ATM trained using types of data: using only the 10 UR videos, using only the 160 Franka videos, and using both Franka and UR videos  (Franka $\Rightarrow$ UR). In the right plot, we observe that the additional cross-embodiment data led to significantly better results. Surprisingly, even if the trajectory model is only trained using Franka videos, it exhibits much better performance than the BC without the Franka videos.}
    \label{fig:rob-rob}
\end{figure*}

\begin{table}[ht]
  \centering
  \caption{Average success rates of human-to-robot experiments. ATM trained with human videos significantly outperforms BC and ATM trained with only 10 robot videos, demonstrating the cross-embodiment capability of ATM.}
  \begin{tabular}{@{}cccccc@{}}
    \toprule
     Method & \specialcell{Teleoperation \\ demos} & \specialcell{Human \\ videos} & \textit{fold cloth} & \textit{put tomato} & \textit{sweep toys} \\
    \midrule
    BC & \CheckmarkBold & \XSolidBrush & 0\% & 10\% & 30\% \\
    ATM & \CheckmarkBold & \XSolidBrush & 0\% & 0\% & 13\% \\
    ATM & \CheckmarkBold & \CheckmarkBold & \textbf{63}\% & \textbf{63\%} & \textbf{60\%} \\
    \bottomrule
  \end{tabular}
  \label{tab:human-robot-result}
\end{table}

\subsection{Human-to-robot and Robot-to-robot Transfer} \label{sec:human_video}
By modeling the low-level any-point trajectories, ATM enables learning from cross-embodiment videos of humans or a different robot performing the task. This facilitates the use of more scalable data sources. To verify this, we present the results of learning from human videos in Fig.~\ref{fig:human_video}, with the quantitative results presented in Table~\ref{tab:human-robot-result}, and the results of cross-robot transfer in Fig.~\ref{fig:rob-rob}. We compare the following methods: \textit{(1)} Learning behavior cloning policies only on the action-labeled data \textit{(2)} ATM, using a track transformer trained only on the limited action-labeled robot data, and (3) ATM, using a track transformer trained on both the action-free human and action-labeled robot data. Experiments show that training the trajectory model on additional cross-embodiment videos makes the trajectory prediction more robust and accurate, significantly improving policy learning. On the other hand, as the number of action-labeled trajectories is small, BC baselines that only use action-labeled trajectories fail. Please refer to the videos on our website for better visualization.

\subsection{Ablation Analysis} \label{sec:ablation}
We conduct a series of ablation experiments in simulation to demonstrate the effect of our design choices. These include the number of action-labeled trajectories needed for policy learning, the effect of the trajectory prediction horizon, image masking, and late track fusion. Besides, we also investigate the universality of ATM for different policy models.

\begin{figure}[ht]
    \centering
    \includegraphics[width=\linewidth]{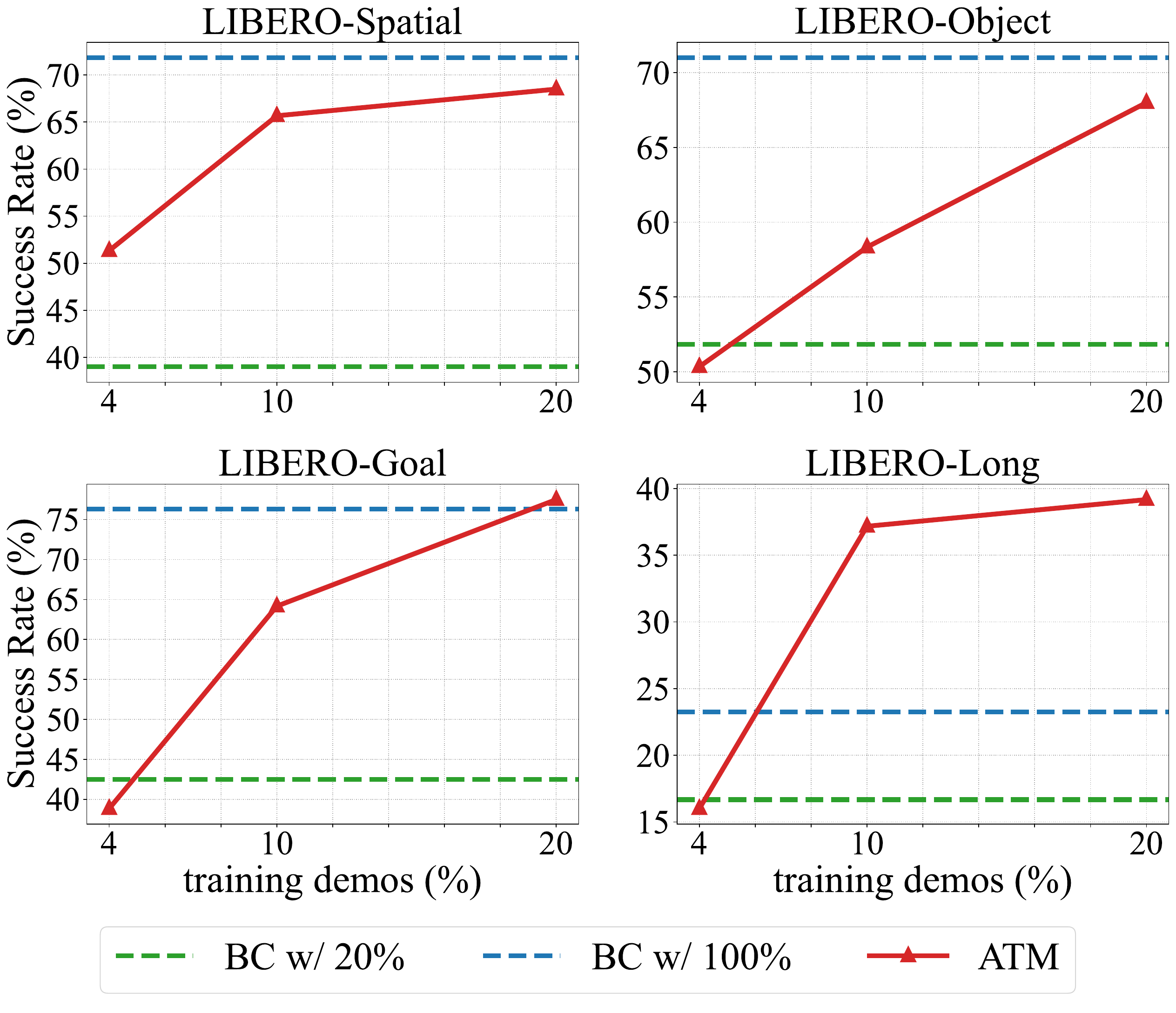}
    \caption{Success rate of our policy trained with 4\%, 10\% and 20\% action-labeled demos. Our policy trained with only 4\% demos performs comparably to BC baseline with 20\% demos on LIBERO-Spatial, Object, and GOAL, and even better on LIBERO-Spatial. When trained on 20\% demos, our performance approaches BC with all training data.}
    \label{fig:num-demos}
\end{figure}

\begin{figure}[ht]
    \centering
    \includegraphics[width=0.9\linewidth]{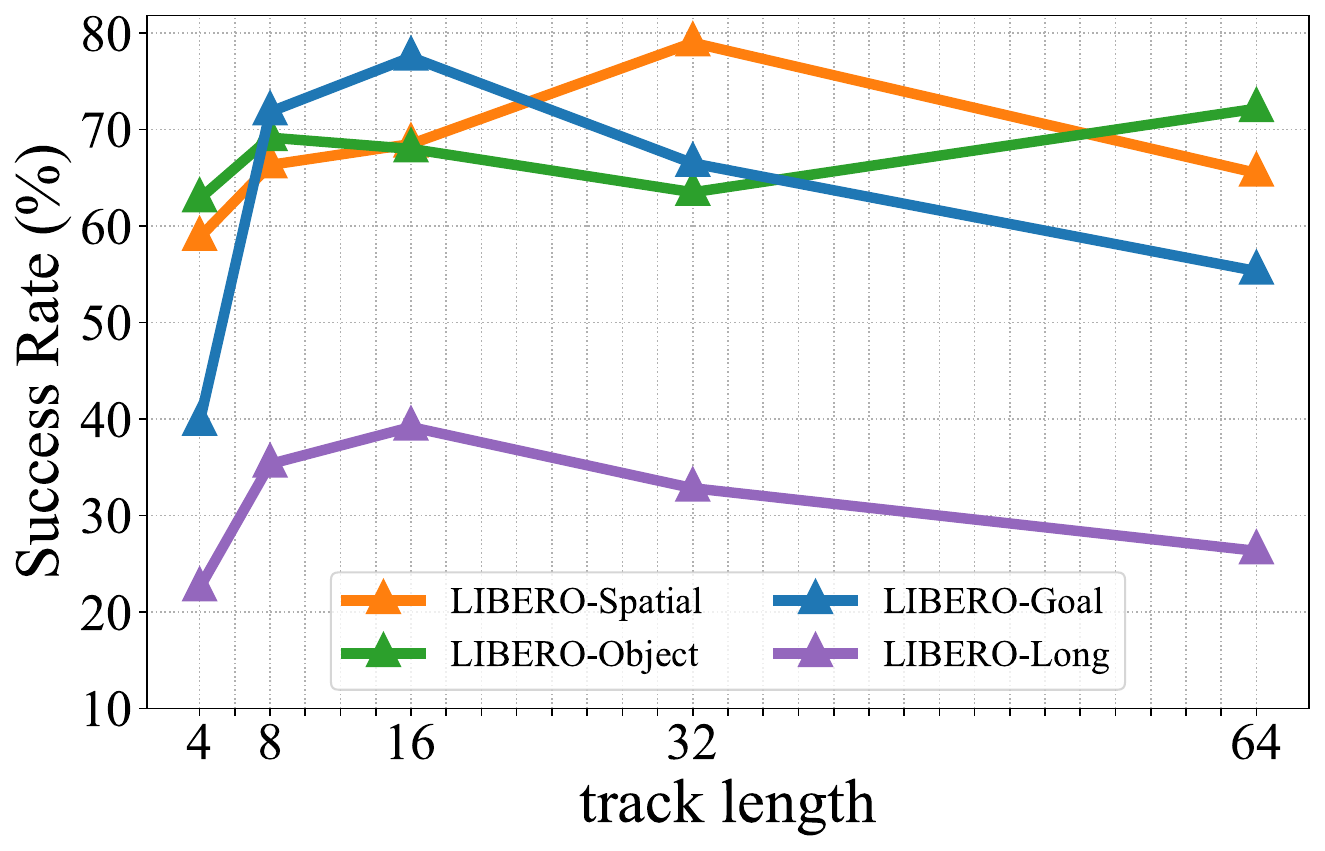}
    \caption{We plot the success rates of the policies learned with predicted trajectories of different lengths. Generally, longer trajectory length improves the performance, but the benefit tends to plateau after 16.}
    \label{fig:libero-track-len}
\end{figure}

\noindent\textbf{Effect of the number of action-labeled trajectories.}
We investigate the impact of track length on our policy input by training our policy using predicted tracks of varying lengths: 4, 8, 16, 32, and 64 steps.
As illustrated in Figure~\ref{fig:libero-track-len}, a track length of 16 yields optimal performance on average. Smaller track lengths significantly decrease performance, while a track length of 0 reduces ATM to the behavior cloning baseline without any benefit of pre-training. On the other hand, larger track lengths can also negatively impact performance, especially on more challenging tasks such as LIBERO-Long and LIBERO-Goal, as longer track predictions are noisier and less relevant for achieving short-horizon subgoals. Determining the optimal time horizon for subgoals could be an interesting avenue for future research.

\begin{table*}[ht]
  \centering
  \caption{Ablation study on image masking of track transformer, where ``w/o image masking" represents that we do not mask out image patches during track transformer training and ``w/ image masking" means we randomly mask 50\% patches. We can see that mask image modeling in track transformer improves the policy performance.}
  \begin{tabular}{@{}ccccc@{}}
    \toprule
    Image Mask Ratio & Spatial & Object & Goal & Long \\
    \midrule
    w/o image masking & $69.17\pm6.38$ & $65.00\pm3.89$ & $74.33\pm3.66$ & $30.83\pm11.43$ \\
    w/ image masking (default) & $68.50\pm1.78$ & $68.00\pm6.18$ & $77.83\pm0.82$ & $39.33\pm15.80$ \\
    \bottomrule
  \end{tabular}
  \label{tab:ablation-mask-ratio}
\end{table*}

\begin{table*}[ht]
  \centering
  \caption{Ablation study on the policy architecture. We explore the effect of the tracks fed into the policy in two positions:  transformer input (early fusion) and MLP head (late fusion), as illustrated in Figure~\ref{fig:track_policy}.}
  \begin{tabular}{@{}cccccc@{}}
    \toprule
    early fusion & late fusion & Spatial & Object & Goal & Long \\
    \midrule
    \CheckmarkBold & \XSolidBrush & $44.67\pm1.84$ & $56.67\pm3.09$ & $5.33\pm0.24$ & $22.33\pm4.94$ \\
    \XSolidBrush & \CheckmarkBold & $65.50\pm3.89$ &  $60.00\pm1.47$ & $72.83\pm4.73$  & $42.76\pm14.62$ \\
    \midrule
    \CheckmarkBold & \CheckmarkBold & $68.50\pm1.78$ & $68.00\pm6.18$ & $77.83\pm0.82$ & $39.33\pm15.80$ \\
    \bottomrule
  \end{tabular}
  \label{tab:arch-ablation}
\end{table*}

\noindent\textbf{Effect of track length.}
The track length of our pre-trained track transformer is 16 and we utilize all 16 points of each track by default as input for our policy.
In this experiment, we investigate the effect of track length of our policy input, by training our policy with the predicted tracks truncated into 4, 8, and 16 steps.
The results in Figure~\ref{fig:libero-track-len} demonstrate that longer tracks result in higher success rates in general, except LIBERO-Object where track length $=8$ yields the best performance. 
In particular, reducing track length leads to larger performance drops on LIBERO-Goal and LIBERO-Long.
These tasks require a comprehensive understanding of the task goal. Longer tracks provide more detailed and specific subgoals for each task, which is crucial for guiding the agent's movements and actions in subsequent stage. 
Conversely, for LIBERO-Object, which emphasizes precision operations over understanding of the task goal, a policy with a length of 16 underperforms slightly compared to the model with a length of 8. We hypothesize that the longer tracks might interfere with the learning of inverse dynamics due to noise. This also supports our approach of employing tracks both as subgoal conditions and in leveraging the nature of inverse dynamics.

\noindent\textbf{Effect of image masking in track transformer.}
When training the track transformer, we randomly mask out patches in the images and learn to reconstruct them as an auxiliary task. We conduct an ablation study by removing the image masking loss and comparing it against our standard configuration, where we randomly mask out $50\%$ of the image masks. In Table~\ref{tab:ablation-mask-ratio}, the results reveal a slight decline in policy performance when image masking is omitted, suggesting image masking can be a useful auxiliary task, except on LIBERO-Spatial. We hypothesize that the auxiliary loss encourages the model to jointly reason about different regions of an image observation. This bias may not be critical for the Libero-spatial tasks because the specific spatial location of interest is already fully specified in the language instruction (e.g., all tasks have the format of "pick up the black bowl at [some spatial location]"). 

\noindent\textbf{Effect of early and late fusion in policy architecture.}
As shown in Figure~\ref{fig:track_policy}, the predicted tracks are fed into the policy both before and after the transformer architecture within the policy, which we call early fusion and late fusion respectively. We conduct ablation studies by removing these two track inputs. The results are shown in Table~\ref{tab:arch-ablation}. We can see that removing the late fusion leads to the most significant performance drop; on LIBERO-Goal, \textit{w/ only early fusion} performs similarly to other baselines, whereas only late fusion performs marginally worse than our full method. This suggests a late fusion of the predicted tracks acts as useful subgoals that help the policy better understand the tasks in a multi-task learning setting. The subgoal prediction is more robust as it is trained on a larger video dataset.

%% file: sec/5_conclusion.tex
\section{Limitations}
One limitation is that our approach still relies on a set of action-labeled demonstration trajectories for mapping to actions, which limits the generalization of the learned policies. Future works can consider learning the trajectory-following policies using reinforcement learning so that no additional demonstration data are needed. Another limitation of our method is that the video dataset we use in this paper only contains small domain gaps. Learning from in-the-wild video dataset poses additional challenges such as multi-modal distribution, diverse camera motions, and sub-optimal motions. We leave these extensions for future work.

\section{Conclusions}
In this work, we present an any-point trajectory modeling framework as video pre-training that effectively learns behaviors and dynamics from action-free video datasets. After pre-training, by learning a track-guided policy, we demonstrate significant improvements over prior state-of-the-art approaches and show effective learning from out-of-distribution human videos.  We show that a particle-based representation is interpretable, structured, and naturally incorporates physical inductive biases such as object permanence. We hope our works will open doors to more exciting directions in learning from videos with structured representations.

\section{Acknowledgement}
This work was supported in part by the BAIR Industrial Consortium, and the InnoHK of the Government of the Hong Kong Special Administrative Region via the Hong Kong Centre for Logistics Robotics.
This work is also supported by the Ministry of Science and Technology of the People's Republic of China, the 2030 Innovation Megaprojects ``Program on New Generation Artificial Intelligence" (Grant No. 2021AAA0150000), and the National Key R\&D Program of China (2022ZD0161700). This work is done when Chuan Wen is visiting UC Berkeley.

%% file: sec/6_supp.tex
\clearpage
\appendices
\section{Additional Experimental Results}

\subsection{Simulation Experiments}

\noindent\textbf{Numerical results.} We report the numeric values for the success rates on LIBERO benchmark in Table~\ref{tab:libero-results}. All methods use 20$\%$ of the demonstration trajectories except the oracle. See Sec.~\ref{sec:app:unipi} for details about the comparisons on UniPi and UniPi-Replan.

\begin{figure*}[ht]
    \centering
    \includegraphics[width=0.95\linewidth]{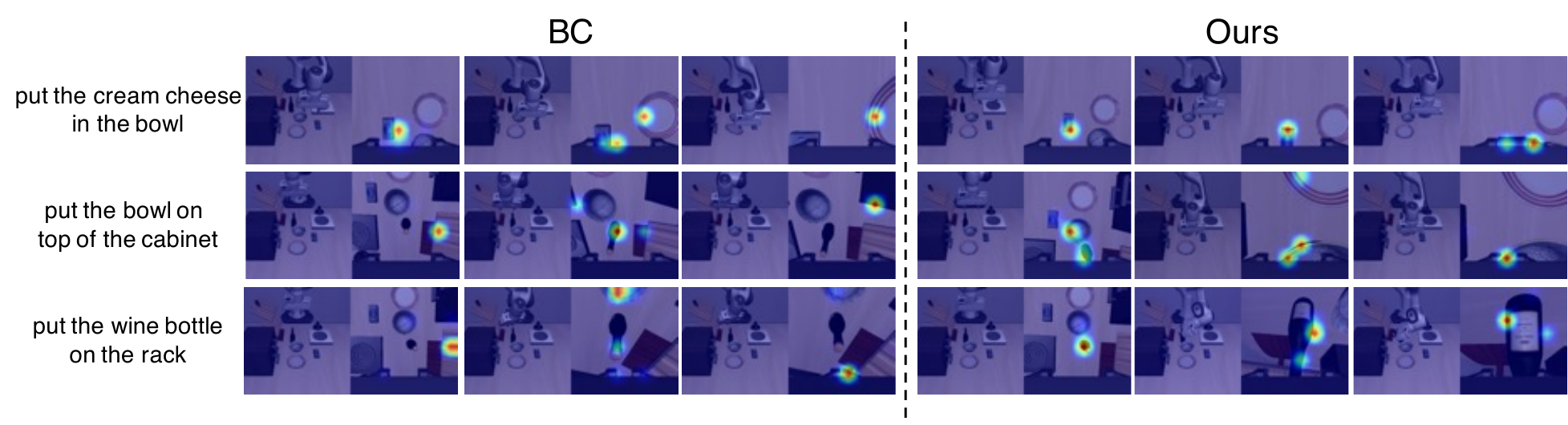}
    \caption{The attention maps of BC and Ours in the spatial transformer. We extract the attention weights between spatial CLS tokens and RGB tokens, highlighting the policy's focus on specific spatial regions during decision-making. The heatmaps reveal our policy's targeted attention on task-relevant areas, in contrast to BC's tendency to focus on irrelevant backgrounds. This underscores the effectiveness of input tracks in the spatial transformer as good task prompts, guiding the CLS token to attend to appropriate areas.}
    \label{fig:policy-attention-map}
\end{figure*}

\noindent\textbf{Attention map visualization of track-guided policy.}
To demonstrate how tracks guide the policy, we visualize the attention maps between the spatial CLS token and RGB tokens in the spatial transformer of BC and our method. Figure~\ref{fig:policy-attention-map} demonstrates that our method effectively focuses on the relevant spatial regions, as specified by the textual instructions. Specifically, it attends to the cream cheese, bowl, and wine bottle in the respective example tasks, while BC is usually distracted by the irrelevant regions, highlighting the superior capability of the tracks as better task prompts.

\subsection{Discussions on the UniPi Baselines}\label{sec:app:unipi}

UniPi~\cite{du2023unipi} proposes to train a language-conditioned video diffusion model $f_\theta$ during video pre-training. During policy learning, given the initial image observation $o_0$ and the language $l$, UniPi first generates all future frames $\Tilde{o}_1, \dots \Tilde{o}_{T-1} = f_\theta (o_0, l)$ and then learns an inverse dynamics model that predicts the action at each time step $a_t = \pi(o_t, \Tilde{o}_{t+1})$. UniPi then executes the actions open-loop. However, training a diffusion model to predict the full video can be computation intensive, As such, the UniPi implementation in our paper follows the one in \citet{Ko2023Learning}, where we predict $N=7$ future frames as the sub-goals for the policy, denoted as $\Tilde{o}_1, \dots \Tilde{o}_N$. During training, we evenly sample $N$ frames in an episode for training the video prediction model. During policy learning, we train a goal-conditioned policy $\pi(a_t, \text{done} | o_t, \Tilde{o}_{i})$, where $i\in {1, \dots N}$ is the image sub-goal and $t$ denotes the current timestep. The policy additionally predicts a done flag to determine when it should switch from the current sub-goal $\Tilde{o}_i$ to the next sub-goal $\Tilde{o}_{i+1}$. ATM's superiority over UniPi can be attributed to two reasons. First, ATM uses a more structured sub-goal representation of point trajectories. Second, ATM performs closed-loop inference, proposing a new sub-goal at each time step, while UniPi's video diffusion process is too slow to be referenced at every time step. 

\begin{table}[ht]
    \centering
    \caption{Computation cost and inference time for different methods on a V100 GPU. ATM performs trajectory generation instead of predicting high-dimensional frames, making it the most computationally efficient and feasible for closed-loop control. UniPi employs a video diffusion model for open-loop future goal generation at the beginning, demanding significantly higher computational resources. UniPi-Replan generates fewer frames than UniPi using a smaller model, resulting in marginally faster generation. However, its use in closed-loop control remains computationally prohibitive.}
    \begin{tabular}{c c c c}
    \toprule
    \multirow{2}{*}{Computation} & \multicolumn{2}{c}{Close-Loop} & Open-Loop \\
     &  ATM & UniPi-Replan & UniPi\\
    \midrule
    TFLOPS per generation        & 1.56 & 13.09 & 39.29 \\
    Time per generation (s)  & 0.015 & 4.51   & 8.14  \\
    \bottomrule
    \end{tabular}
    \label{tab:runtime}
\end{table}

To train the UniPi policy, we sample $o_{i}$ from a future frame $o_{t}$ where $t \in [t, t+t_{max}]$. We choose $t_{max}$ for each task suite ($t_{max} = 16$ for LIBERO-Object and LIBERO-Spatial, $t_{max} = 50$ for LIBERO-Goal and LIBERO-Long). To mitigate dataset imbalance when learning the done flag, we sample $o_{i}$ as the next frame 10\% of the time. We perform MSE regression on both the action $a_t$ and done.

In order to decouple the two advantages of ATM over UniPi additionally compare with a UniPi variation where we train the video prediction model to predict a fixed time step into the future $\Tilde{o}_{t+H}= f_\theta(o_t)$ for every $H$ steps of policy execution, where $H=8$. As this variation replans the sub-goal more frequently, we call this method \textit{UniPi-Replan}, similar to the implementation in~\cite{black2023zero}. The results are shown in Table~\ref{tab:libero-results}. Surprisingly, we found that this variation performs even worse than UniPi. \textbf{We thus draw the conclusion that a structured sub-goal representation can be much more effective than an image sub-goal.}  The reason is that predicting an image goal at a fixed future time step can be a difficult objective for the video prediction model, leading to inconsistent subgoals. Please refer to the failure videos of various baselines on our website. Additionally, this method requires heavy computation. A comparison of the computation needed is shown in Table~\ref{tab:runtime}. Due to the computation cost, we only evaluate UniPi-Replan on LIBERO-Object and LIBERO-Goal and report the average success across 10 trials on one policy training random seed.

\begin{table*}[ht]
  \centering
  \caption{Average success rate on LIBERO benchmark. Our method performs consistently better than all the baselines across all suites. UniPi-Replan is only evaluated for a single seed due to the computation cost.}
  \begin{tabular}{@{}ccccccc@{}}
    \toprule
    Method & Libero-Spatial & Libero-Object & Libero-Goal & Libero-Long & Libero-90 \\
    \midrule
    BC-Full-Trainset (Oracle) & $71.83\pm3.70$ & $71.00\pm7.97$ & $76.33\pm1.31$ & $24.17\pm2.59$ & - \\
    \midrule
    BC & $39.00\pm8.20$ & $51.83\pm13.54$ & $42.50\pm4.95$ & $16.67\pm3.66$ & $29.78\pm1.14$ \\
    R3M-finetune~\cite{nair2023r3m} & $49.17\pm3.79$ & $52.83\pm2.25$ & $5.33\pm1.43$ & $9.17\pm2.66$ & $9.59\pm0.27$ \\
    VPT~\cite{baker2022vpt} & $37.83\pm4.29$ & $19.50\pm0.82$ & $3.33\pm2.36$ & $3.83\pm1.65$ & - \\
    UniPi~\cite{du2023unipi,Ko2023Learning} & $\bf 69.17 \pm 3.75$ & $59.83 \pm 3.01$ & $11.83 \pm 2.02$ & $5.83 \pm 2.08$ & - \\
    UniPi-Replan~\cite{black2023zero} & - & 31.00 & 3.00 & - & - \\
    ATM (Ours) & $68.50\pm1.78$ & $\bf 68.00\pm6.18$ & $\bf 77.83\pm0.82$ & $\bf 39.33\pm15.80$ & $\bf 48.41\pm2.09$ \\
    \bottomrule
  \end{tabular}
  \label{tab:libero-results}
\end{table*}

\begin{figure}[ht]
  \centering
  \includegraphics[width=0.9\linewidth]{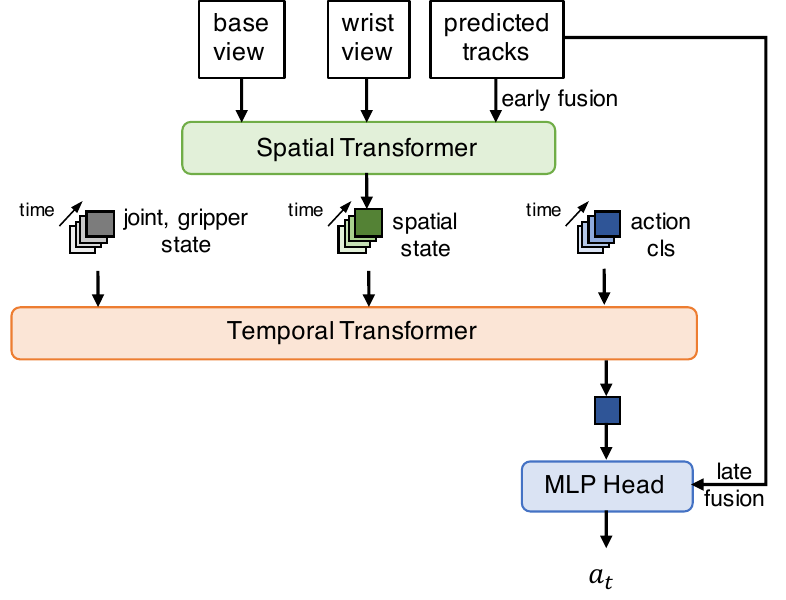}
  \caption{To summarize spatial information, we perform self-attention on a sequence consisting of all views' track and image patches and a CLS token. To integrate information across time, we perform casual self-attention between spatial CLS token, proprioception, and an action CLS token per timestep. To regress actions, we concatenate each timestep's action CLS token and proposed tracks. }
  \label{fig:arch-appendix}
\end{figure}

\begin{figure}[ht]
  \centering
  \includegraphics[width=\linewidth]{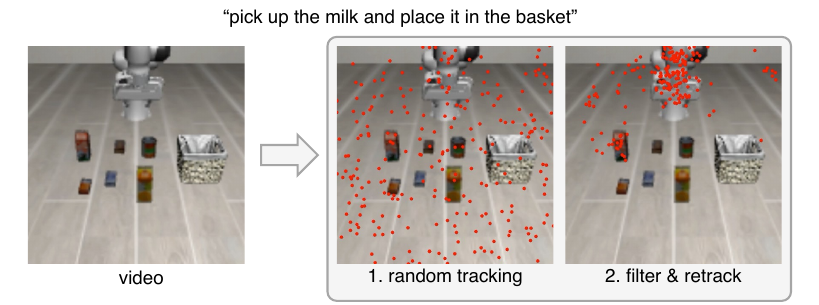}
  \caption{Given a video (left), we query 1000 randomly sampled points using an off-the-shelf TAP model (middle), where each colored dot represents the starting position of a track. We then filter the tracks using a heuristic of their position displacement across the video and re-sample around these points (right). We can see that extracted tracks are concentrated around informative objects, such as the robot's gripper and manipulation targets.}
  \label{fig:filtering}
\end{figure}

\subsection{Applying ATM to Diffusion Policy}
In addition to the transformer policy adopted from the original LIBERO paper~\cite{liu2023libero}, our ATM can also be integrated with the state-of-the-art Diffusion Policy~\cite{Chi2023diffusion}. Specifically, we implement an ATM Diffusion Policy by using predicted tracks as input conditions. Both the standard and ATM-based Diffusion policies are trained with 20\% of the data from LIBERO, in the same setting as the experiment in Figure~\ref{fig:sim_exp}.

As shown in Table~\ref{tab:diff-policy}, the standard Diffusion Policy achieves impressive results on the LIBERO benchmark, even with limited training data, confirming its status as the state-of-the-art imitation policy. Moreover, the ATM Diffusion Policy consistently outperforms the standard model, highlighting our framework's versatility and effectiveness across different policy architectures.

\begin{table*}[ht]
  \centering
  \caption{Detailed results of Diffusion policy on LIBERO. The Diffusion policy can be further improved by our method, suggesting that our Any-point Trajectory Modeling framework is an important building block to apply to any policy model.}
  \begin{tabular}{@{}cccccc@{}}
    \toprule
    Method & Libero-Spatial & Libero-Object & Libero-Goal & Libero-Long & Libero-90 \\
    \midrule
    Diffusion Policy & $67.67\pm1.25$ & $78.00\pm2.45$ & $35.00\pm3.74$ & $37.33\pm2.05$  & $33.85\pm1.71$ \\
    ATM Diffusion Policy & $79.00\pm3.74$ & $81.00\pm2.45$ & $58.67\pm4.64$ & $44.00\pm6.38$ & $62.89\pm1.10$ \\
    \bottomrule
  \end{tabular}
  \label{tab:diff-policy}
\end{table*}

\subsection{Human-to-robot Transfer Details}
To demonstrate the potential of ATM to leverage out-of-domain videos, we design human-to-robot transfer tasks in three different settings: 1) deformable object: \textit{fold the cloth and pull it to the right}, 2) long horizon: \textit{put the tomato into the pan and close the cabinet door}, and 3) tool using: \textit{use the broom to sweep the toys into the dustpan and put it in front of the dustpan}.
We collect 10 robot teleportation trajectories (action-labeled) and 100 human manipulation videos (action-free). We compare three methods: 1) behavioral cloning only with 10 action-labeled robot demos, 2) ATM, training track transformer with only 10 robot demos, and 3) ATM, training a track transformer with both action-free human and action-labeled robot data.
For each task, we train each method with three different random seeds, evaluate them in the real world 10 times, and report the average success rate in Table~\ref{tab:human-robot-result}.

\begin{figure*}[ht]
    \centering
    \includegraphics[width=\linewidth]{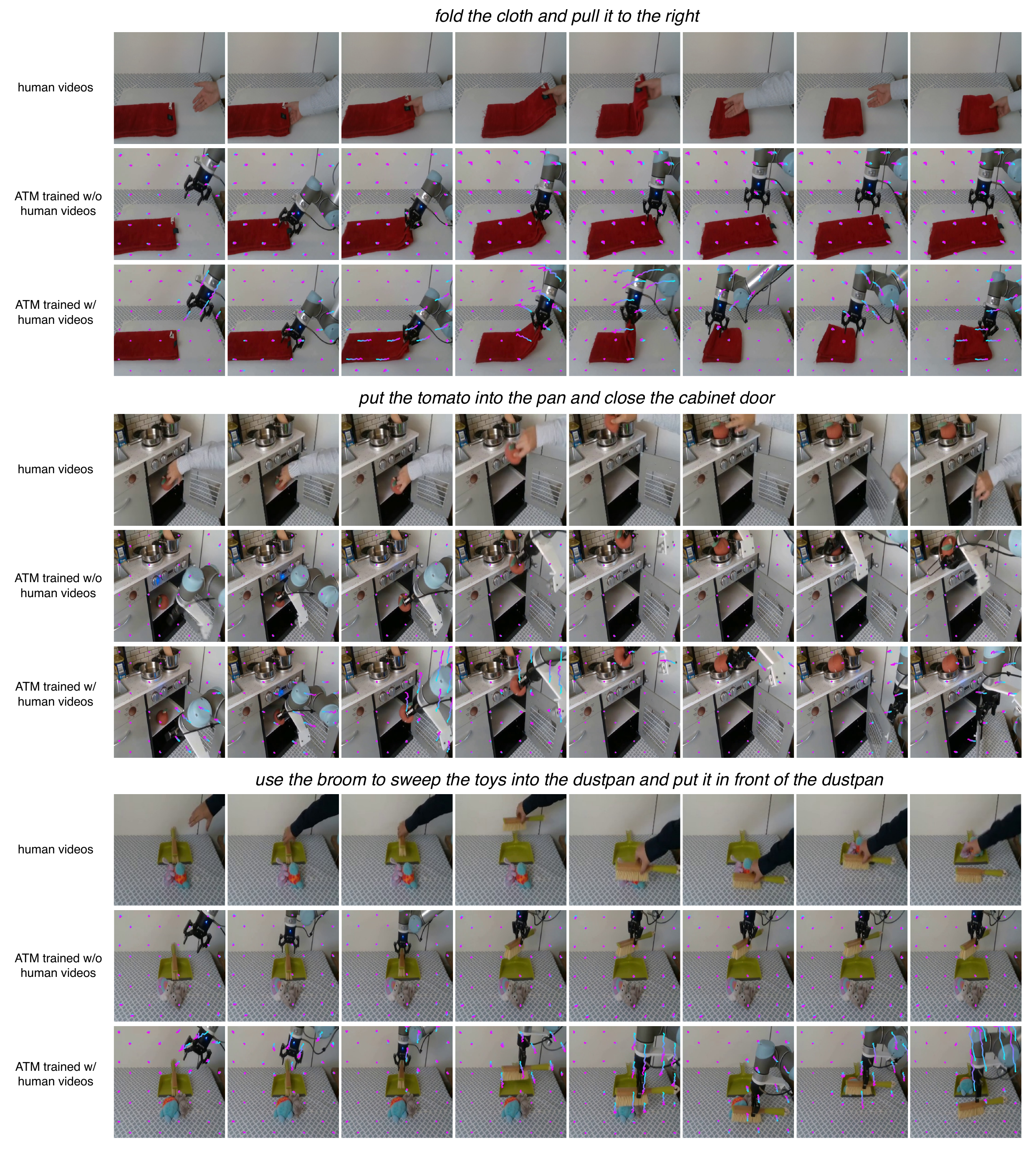}
    \caption{The visualizations of human demos and rollout videos of ATM policies trained with and without human data. We can see that ATM is able to take advantage of out-of-domain videos, i.e., human videos, to generate more precise tracks, resulting in better policy performance.}
    \label{fig:human_video_supp}
\end{figure*}

\noindent\textbf{Human-to-Robot visualization.}
The detailed video visualizations of human-to-robot skill transfer are shown in Figure~\ref{fig:human_video_supp}. Due to limited training samples, the track transformer trained with only 10 robot videos fails to predict the future point trajectories, leading to low success rates. In contrast, the trajectory model trained with human videos generates high-quality tracks in each frame, guiding the agent to successfully complete the task with only 10 action-labeled trajectories. This indicates our any-point trajectory modeling enables to transfer the motion prior from cross-embodiment videos to robot skills, which significantly improves policy learning.

\begin{figure*}
    \centering
    \includegraphics[width=0.95\linewidth]{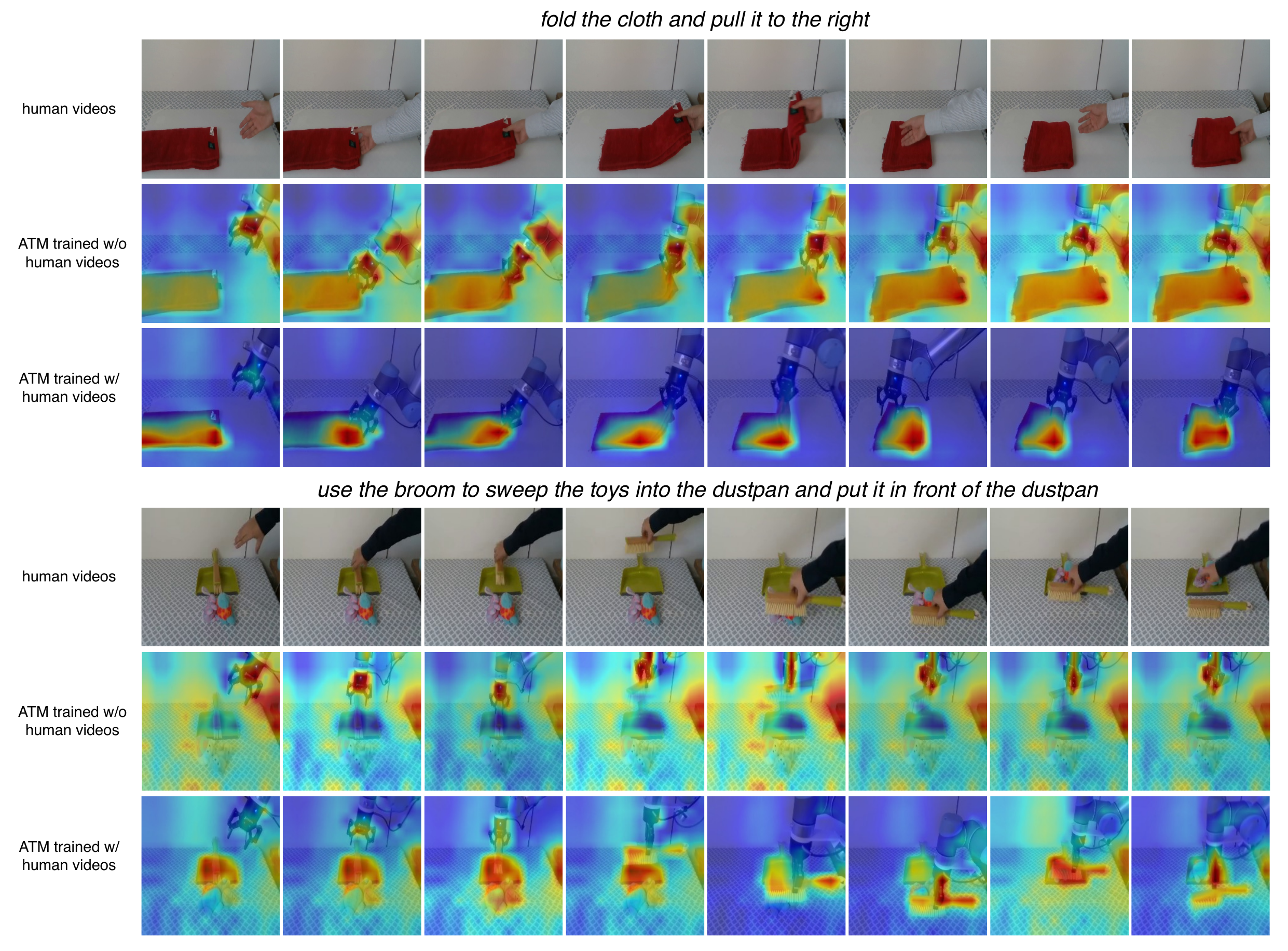}
    \caption{Attention maps of Track Transformer trained with and without human videos. Including large-scale human video leads to much clearer attention maps, focusing on the object and robot arm, while training without human videos will attend to incorrect areas such as background walls.}
    \label{fig:tt-attention-map}
\end{figure*}

Furthermore, we visualize the attention maps of Track Transformers trained with or without human videos in Figure~\ref{fig:tt-attention-map}. Track Transformer trained with human videos (bottom row) displays well-defined attention heatmaps that primarily focus on the object and robot arm. Conversely, the attention maps from the model trained without human videos (middle row) are notably more dispersed, often incorrectly focusing on background areas. This comparison highlights the effectiveness of our ATM in leveraging prior knowledge from human videos, facilitating successful cross-embodiment imitation learning.

\begin{table*}[ht]
\begin{minipage}[c]{0.48\textwidth}
\makeatletter\def\@captype{table}
\centering
\caption{Hyperparameters of track transformer training.}
\begin{tabular}{@{}cc@{}}
    \toprule
    Hyperparameters & Track Transformer \\
    \midrule
    epoch & 100  \\
    batch size &  1024  \\
    optimizer & AdamW  \\
    learning rate & 1e-4  \\
    weight decay & 1e-4 \\
    lr scheduler & Cosine  \\\
    lr warm up & 5 \\
    clip grad & 10 \\
    point sampling & variance filtering \\
    number of points & 32 \\
    track length & 16 \\
    track patch size & 4 \\
    image mask ratio & 0.5 \\
    augmentation & ColorJitter,RandomShift \\
    \bottomrule
\end{tabular}
\label{tab:tt-hyperparameter}
\end{minipage}
\begin{minipage}[c]{0.48\textwidth}
\makeatletter\def\@captype{table}
\centering
  \caption{Hyperparameters of policy training.}
  \begin{tabular}{@{}cc@{}}
    \toprule
    Hyperparameters & Policy  \\
    \midrule
    epoch & 100  \\
    batch size &  512  \\
    optimizer &  AdamW  \\
    learning rate & 5e-4  \\
    weight decay & 1e-4 \\
    lr scheduler & Cosine  \\
    lr warm up & 0 \\
    clip grad & 100 \\
    point sampling & grid \\
    number of points & 32 \\
    track length & 16 \\
    frame stack & 10 \\
    augmentation & ColorJitter,RandomShift \\
    \bottomrule
  \end{tabular}
  \label{tab:policy-hyperparameter}
\end{minipage}
\end{table*}

\section{Implementation Details}
\subsection{Policy Architecture}~\label{sec:app:policy-arch}
We include a more detailed visual of the ViT-T~\cite{liu2023libero, kim2021vilt} policy used in our experiments in Figure~\ref{fig:arch-appendix}. The input to our policy is a set of temporally stacked images across multiple views $o_t \in \mathbb{R}^{V\times T \times C\times H\times W}$, and proprioception $p_t \in \mathbb{R}^{D_p}$. To process these inputs, ViT-T consists of three stages: \\

\noindent\textbf{Spatial Encoding}: We encode all modalities at each timestep. We first leverage the frozen track transformer to propose a set of tracks for all $V$ frames in $o_t$. We then project each modality with modality-specific encoders into a shared embedding space $\mathbb{R}^D$. Each modality's tokens are embedded using a shared learned modality token, and modality-specific positional embeddings. We then concatenate tokens across all modalities and views with a learned spatial CLS token, and perform self-attention on the sequence. We extract the spatial CLS token as the representation. \\

\noindent\textbf{Temporal Decoding}: We process the encoded modalities across timesteps into actions. We first project the proprioception at each timestep into the same shared embedding space $\mathbb{R}^D$. We then interleave the encoded proprioception, the spatial CLS token, and a learned action CLS token across timesteps into a sequence, before performing causally-masked self attention between the sequence. \\

\noindent\textbf{Action Head}: We treat each timestep's output independently and parameterize actions using an MLP. For each timestep, we take the action CLS token, and fuse the CLS token with the reconstructed tracks of the current timestep. \\

\subsection{Efficient Training with Point Filtering}\label{sec:app:track-filter}
We adopt a heuristic to filter out the static points in the background and then utilize an off-the-shelf tracking model to generate the corresponding tracks of the large-motion points. The visualization of the sampled points before and after the filtering process is shown in Figure~\ref{fig:filtering}.

\subsection{Training Details}~\label{sec:app:train-detail}
We list the training hyperparameters for the track transformer and track-guided policy in Table~\ref{tab:tt-hyperparameter} and \ref{tab:policy-hyperparameter}, which are fixed for all experiments on LIBERO benchmark. We train all models on 4 A100 GPUs with DeepSpeed strategy.

We train the track transformer using the ground truth tracks generated by CoTracker~\cite{karaev2023cotracker} and save the checkpoint with the lowest validation loss as our final model to apply for policy learning. We do not incorporate frame stacking for track transformer to avoid causal confusion~\cite{wen2022primenet}. 

We train policies using the expert demonstrations provided by LIBERO, which is collected by human experts through teleoperation with 3Dconnexion Spacemouse~\cite{liu2023libero}. Since the validation loss in behavioral cloning is not always reliable, we save the checkpoint of the last epoch for online rollout evaluation.

%% file: main.bbl
\begin{thebibliography}{57}
\providecommand{\natexlab}[1]{#1}
\providecommand{\url}[1]{\texttt{#1}}
\expandafter\ifx\csname urlstyle\endcsname\relax
  \providecommand{\doi}[1]{doi: #1}\else
  \providecommand{\doi}{doi: \begingroup \urlstyle{rm}\Url}\fi

\bibitem[Aguiar and Hespanha(2007)]{aguiar2007trajectory}
A~Pedro Aguiar and Joao~P Hespanha.
\newblock Trajectory-tracking and path-following of underactuated autonomous vehicles with parametric modeling uncertainty.
\newblock \emph{IEEE transactions on automatic control}, 52\penalty0 (8):\penalty0 1362--1379, 2007.

\bibitem[Bahl et~al.(2023)Bahl, Mendonca, Chen, Jain, and Pathak]{bahl418affordances}
S~Bahl, R~Mendonca, L~Chen, U~Jain, and D~Pathak.
\newblock Affordances from human videos 417 as a versatile representation for robotics.
\newblock In \emph{CVPR}, 2023.

\bibitem[Baker et~al.(2022)Baker, Akkaya, Zhokov, Huizinga, Tang, Ecoffet, Houghton, Sampedro, and Clune]{baker2022vpt}
Bowen Baker, Ilge Akkaya, Peter Zhokov, Joost Huizinga, Jie Tang, Adrien Ecoffet, Brandon Houghton, Raul Sampedro, and Jeff Clune.
\newblock Video pretraining (vpt): Learning to act by watching unlabeled online videos.
\newblock \emph{Advances in Neural Information Processing Systems}, 35:\penalty0 24639--24654, 2022.

\bibitem[Bharadhwaj et~al.(2023)Bharadhwaj, Gupta, Tulsiani, and Kumar]{bharadhwaj2023zero}
Homanga Bharadhwaj, Abhinav Gupta, Shubham Tulsiani, and Vikash Kumar.
\newblock Zero-shot robot manipulation from passive human videos.
\newblock \emph{arXiv preprint arXiv:2302.02011}, 2023.

\bibitem[Black et~al.(2023)Black, Nakamoto, Atreya, Walke, Finn, Kumar, and Levine]{black2023zero}
Kevin Black, Mitsuhiko Nakamoto, Pranav Atreya, Homer Walke, Chelsea Finn, Aviral Kumar, and Sergey Levine.
\newblock Zero-shot robotic manipulation with pretrained image-editing diffusion models.
\newblock \emph{arXiv preprint arXiv:2310.10639}, 2023.

\bibitem[Brohan et~al.(2022)Brohan, Brown, Carbajal, Chebotar, Dabis, Finn, Gopalakrishnan, Hausman, Herzog, Hsu, et~al.]{brohan2022rt}
Anthony Brohan, Noah Brown, Justice Carbajal, Yevgen Chebotar, Joseph Dabis, Chelsea Finn, Keerthana Gopalakrishnan, Karol Hausman, Alex Herzog, Jasmine Hsu, et~al.
\newblock Rt-1: Robotics transformer for real-world control at scale.
\newblock \emph{arXiv preprint arXiv:2212.06817}, 2022.

\bibitem[Brown et~al.(2020)Brown, Mann, Ryder, Subbiah, Kaplan, Dhariwal, Neelakantan, Shyam, Sastry, Askell, et~al.]{brown2020language}
Tom Brown, Benjamin Mann, Nick Ryder, Melanie Subbiah, Jared~D Kaplan, Prafulla Dhariwal, Arvind Neelakantan, Pranav Shyam, Girish Sastry, Amanda Askell, et~al.
\newblock Language models are few-shot learners.
\newblock \emph{Advances in neural information processing systems}, 33:\penalty0 1877--1901, 2020.

\bibitem[Chi et~al.(2023)Chi, Feng, Du, Xu, Cousineau, Burchfiel, and Song]{Chi2023diffusion}
Cheng Chi, Siyuan Feng, Yilun Du, Zhenjia Xu, Eric Cousineau, Benjamin~CM Burchfiel, and Shuran Song.
\newblock {Diffusion Policy: Visuomotor Policy Learning via Action Diffusion}.
\newblock In \emph{Proceedings of Robotics: Science and Systems}, Daegu, Republic of Korea, July 2023.
\newblock \doi{10.15607/RSS.2023.XIX.026}.

\bibitem[Devlin et~al.(2019)Devlin, Chang, Lee, and Toutanova]{devlin2019-bert}
Jacob Devlin, Ming-Wei Chang, Kenton Lee, and Kristina Toutanova.
\newblock {BERT}: Pre-training of deep bidirectional transformers for language understanding.
\newblock In Jill Burstein, Christy Doran, and Thamar Solorio, editors, \emph{Proceedings of the 2019 Conference of the North {A}merican Chapter of the Association for Computational Linguistics: Human Language Technologies, Volume 1 (Long and Short Papers)}, pages 4171--4186, Minneapolis, Minnesota, June 2019. Association for Computational Linguistics.

\bibitem[Doersch et~al.(2023{\natexlab{a}})Doersch, Gupta, Markeeva, Recasens, Smaira, Aytar, Carreira, Zisserman, and Yang]{doersch2023tapvid}
Carl Doersch, Ankush Gupta, Larisa Markeeva, Adrià Recasens, Lucas Smaira, Yusuf Aytar, João Carreira, Andrew Zisserman, and Yi~Yang.
\newblock Tap-vid: A benchmark for tracking any point in a video, 2023{\natexlab{a}}.

\bibitem[Doersch et~al.(2023{\natexlab{b}})Doersch, Yang, Vecerik, Gokay, Gupta, Aytar, Carreira, and Zisserman]{doersch2023tapir}
Carl Doersch, Yi~Yang, Mel Vecerik, Dilara Gokay, Ankush Gupta, Yusuf Aytar, Joao Carreira, and Andrew Zisserman.
\newblock Tapir: Tracking any point with per-frame initialization and temporal refinement, 2023{\natexlab{b}}.

\bibitem[Du et~al.(2023)Du, Yang, Dai, Dai, Nachum, Tenenbaum, Schuurmans, and Abbeel]{du2023unipi}
Yilun Du, Sherry Yang, Bo~Dai, Hanjun Dai, Ofir Nachum, Joshua~B Tenenbaum, Dale Schuurmans, and Pieter Abbeel.
\newblock Learning universal policies via text-guided video generation.
\newblock In \emph{Thirty-seventh Conference on Neural Information Processing Systems}, 2023.

\bibitem[Escontrela et~al.(2023)Escontrela, Adeniji, Yan, Jain, Peng, Goldberg, Lee, Hafner, and Abbeel]{Escontrela23arXiv_VIPER}
Alejandro Escontrela, Ademi Adeniji, Wilson Yan, Ajay Jain, Xue~Bin Peng, Ken Goldberg, Youngwoon Lee, Danijar Hafner, and Pieter Abbeel.
\newblock Video prediction models as rewards for reinforcement learning.
\newblock \emph{Neural Information Processing Systems}, 2023.

\bibitem[Fang et~al.(2023)Fang, Fang, Tang, Liu, Wang, Wang, Zhu, and Lu]{fang2023rh20t}
Hao-Shu Fang, Hongjie Fang, Zhenyu Tang, Jirong Liu, Chenxi Wang, Junbo Wang, Haoyi Zhu, and Cewu Lu.
\newblock Rh20t: A comprehensive robotic dataset for learning diverse skills in one-shot.
\newblock In \emph{Towards Generalist Robots: Learning Paradigms for Scalable Skill Acquisition@ CoRL2023}, 2023.

\bibitem[Goyal et~al.(2022)Goyal, Mousavian, Paxton, Chao, Okorn, Deng, and Fox]{goyal2022ifor}
Ankit Goyal, Arsalan Mousavian, Chris Paxton, Yu-Wei Chao, Brian Okorn, Jia Deng, and Dieter Fox.
\newblock Ifor: Iterative flow minimization for robotic object rearrangement.
\newblock In \emph{Proceedings of the IEEE/CVF Conference on Computer Vision and Pattern Recognition}, pages 14787--14797, 2022.

\bibitem[Gu et~al.(2023)Gu, Kirmani, Wohlhart, Lu, Arenas, Rao, Yu, Fu, Gopalakrishnan, Xu, et~al.]{gu2023rt}
Jiayuan Gu, Sean Kirmani, Paul Wohlhart, Yao Lu, Montserrat~Gonzalez Arenas, Kanishka Rao, Wenhao Yu, Chuyuan Fu, Keerthana Gopalakrishnan, Zhuo Xu, et~al.
\newblock Rt-trajectory: Robotic task generalization via hindsight trajectory sketches.
\newblock \emph{arXiv preprint arXiv:2311.01977}, 2023.

\bibitem[Harley et~al.(2022)Harley, Fang, and Fragkiadaki]{harley2022particle}
Adam~W Harley, Zhaoyuan Fang, and Katerina Fragkiadaki.
\newblock Particle video revisited: Tracking through occlusions using point trajectories.
\newblock In \emph{European Conference on Computer Vision}, pages 59--75. Springer, 2022.

\bibitem[He et~al.(2022)He, Chen, Xie, Li, Doll{\'a}r, and Girshick]{he2022masked}
Kaiming He, Xinlei Chen, Saining Xie, Yanghao Li, Piotr Doll{\'a}r, and Ross Girshick.
\newblock Masked autoencoders are scalable vision learners.
\newblock In \emph{Proceedings of the IEEE/CVF conference on computer vision and pattern recognition}, pages 16000--16009, 2022.

\bibitem[Huang et~al.(2022)Huang, Lin, and Held]{huang2022medor}
Zixuan Huang, Xingyu Lin, and David Held.
\newblock Mesh-based dynamics model with occlusion reasoning for cloth manipulation.
\newblock In \emph{Robotics: Science and Systems (RSS)}, 2022.

\bibitem[Karaev et~al.(2023)Karaev, Rocco, Graham, Neverova, Vedaldi, and Rupprecht]{karaev2023cotracker}
Nikita Karaev, Ignacio Rocco, Benjamin Graham, Natalia Neverova, Andrea Vedaldi, and Christian Rupprecht.
\newblock Cotracker: It is better to track together.
\newblock \emph{arXiv:2307.07635}, 2023.

\bibitem[Kim et~al.(2021)Kim, Son, and Kim]{kim2021vilt}
Wonjae Kim, Bokyung Son, and Ildoo Kim.
\newblock Vilt: Vision-and-language transformer without convolution or region supervision, 2021.

\bibitem[Kirillov et~al.(2023)Kirillov, Mintun, Ravi, Mao, Rolland, Gustafson, Xiao, Whitehead, Berg, Lo, Dollar, and Girshick]{kirillov2023segment}
Alexander Kirillov, Eric Mintun, Nikhila Ravi, Hanzi Mao, Chloe Rolland, Laura Gustafson, Tete Xiao, Spencer Whitehead, Alexander~C. Berg, Wan-Yen Lo, Piotr Dollar, and Ross Girshick.
\newblock Segment anything.
\newblock In \emph{Proceedings of the IEEE/CVF International Conference on Computer Vision (ICCV)}, pages 4015--4026, October 2023.

\bibitem[Ko et~al.(2023)Ko, Mao, Du, Sun, and Tenenbaum]{Ko2023Learning}
Po-Chen Ko, Jiayuan Mao, Yilun Du, Shao-Hua Sun, and Joshua~B Tenenbaum.
\newblock {Learning to Act from Actionless Video through Dense Correspondences}.
\newblock \emph{arXiv:2310.08576}, 2023.

\bibitem[Li et~al.(2021)Li, Li, Sitzmann, Agrawal, and Torralba]{li20213d}
Yunzhu Li, Shuang Li, Vincent Sitzmann, Pulkit Agrawal, and Antonio Torralba.
\newblock 3d neural scene representations for visuomotor control.
\newblock \emph{arXiv preprint arXiv:2107.04004}, 2021.

\bibitem[Lin et~al.(2021)Lin, Wang, Huang, and Held]{lin2021VCD}
Xingyu Lin, Yufei Wang, Zixuan Huang, and David Held.
\newblock Learning visible connectivity dynamics for cloth smoothing.
\newblock In \emph{Conference on Robot Learning}, 2021.

\bibitem[Lin et~al.(2023)Lin, So, Mahalingam, Liu, and Abbeel]{lin2023spawnnet}
Xingyu Lin, John So, Sashwat Mahalingam, Fangchen Liu, and Pieter Abbeel.
\newblock Spawnnet: Learning generalizable visuomotor skills from pre-trained networks.
\newblock \emph{arXiv preprint arXiv:2307.03567}, 2023.

\bibitem[Liu et~al.(2023)Liu, Zhu, Gao, Feng, Zhu, Stone, et~al.]{liu2023libero}
Bo~Liu, Yifeng Zhu, Chongkai Gao, Yihao Feng, Yuke Zhu, Peter Stone, et~al.
\newblock Libero: Benchmarking knowledge transfer for lifelong robot learning.
\newblock In \emph{Thirty-seventh Conference on Neural Information Processing Systems Datasets and Benchmarks Track}, 2023.

\bibitem[Ma et~al.(2023)Ma, Sodhani, Jayaraman, Bastani, Kumar, and Zhang]{ma2023vip}
Yecheng~Jason Ma, Shagun Sodhani, Dinesh Jayaraman, Osbert Bastani, Vikash Kumar, and Amy Zhang.
\newblock {VIP}: Towards universal visual reward and representation via value-implicit pre-training.
\newblock In \emph{The Eleventh International Conference on Learning Representations}, 2023.

\bibitem[Mandlekar et~al.(2021{\natexlab{a}})Mandlekar, Xu, Wong, Nasiriany, Wang, Kulkarni, Fei-Fei, Savarese, Zhu, and Mart{\'\i}n-Mart{\'\i}n]{mandlekar2021matters}
Ajay Mandlekar, Danfei Xu, Josiah Wong, Soroush Nasiriany, Chen Wang, Rohun Kulkarni, Li~Fei-Fei, Silvio Savarese, Yuke Zhu, and Roberto Mart{\'\i}n-Mart{\'\i}n.
\newblock What matters in learning from offline human demonstrations for robot manipulation.
\newblock \emph{arXiv preprint arXiv:2108.03298}, 2021{\natexlab{a}}.

\bibitem[Mandlekar et~al.(2021{\natexlab{b}})Mandlekar, Xu, Wong, Nasiriany, Wang, Kulkarni, Fei-Fei, Savarese, Zhu, and Mart\'{i}n-Mart\'{i}n]{robomimic2021}
Ajay Mandlekar, Danfei Xu, Josiah Wong, Soroush Nasiriany, Chen Wang, Rohun Kulkarni, Li~Fei-Fei, Silvio Savarese, Yuke Zhu, and Roberto Mart\'{i}n-Mart\'{i}n.
\newblock What matters in learning from offline human demonstrations for robot manipulation.
\newblock In \emph{arXiv preprint arXiv:2108.03298}, 2021{\natexlab{b}}.

\bibitem[Manuelli et~al.(2020)Manuelli, Li, Florence, and Tedrake]{manuelli2020keypoints}
Lucas Manuelli, Yunzhu Li, Pete Florence, and Russ Tedrake.
\newblock Keypoints into the future: Self-supervised correspondence in model-based reinforcement learning, 2020.

\bibitem[Mendonca et~al.(2023)Mendonca, Bahl, and Pathak]{mendonca2023structured}
Russell Mendonca, Shikhar Bahl, and Deepak Pathak.
\newblock Structured world models from human videos.
\newblock In \emph{RSS}, 2023.

\bibitem[Nair et~al.(2023)Nair, Rajeswaran, Kumar, Finn, and Gupta]{nair2023r3m}
Suraj Nair, Aravind Rajeswaran, Vikash Kumar, Chelsea Finn, and Abhinav Gupta.
\newblock R3m: A universal visual representation for robot manipulation.
\newblock In \emph{Conference on Robot Learning}, pages 892--909. PMLR, 2023.

\bibitem[Padalkar et~al.(2023)Padalkar, Pooley, Jain, Bewley, Herzog, Irpan, Khazatsky, Rai, Singh, Brohan, et~al.]{padalkar2023open}
Abhishek Padalkar, Acorn Pooley, Ajinkya Jain, Alex Bewley, Alex Herzog, Alex Irpan, Alexander Khazatsky, Anant Rai, Anikait Singh, Anthony Brohan, et~al.
\newblock Open x-embodiment: Robotic learning datasets and rt-x models.
\newblock \emph{arXiv preprint arXiv:2310.08864}, 2023.

\bibitem[Peng et~al.(2018)Peng, Abbeel, Levine, and Van~de Panne]{peng2018deepmimic}
Xue~Bin Peng, Pieter Abbeel, Sergey Levine, and Michiel Van~de Panne.
\newblock Deepmimic: Example-guided deep reinforcement learning of physics-based character skills.
\newblock \emph{ACM Transactions On Graphics (TOG)}, 37\penalty0 (4):\penalty0 1--14, 2018.

\bibitem[Qin et~al.(2020)Qin, Fang, Zhu, Fei-Fei, and Savarese]{qin2020keto}
Zengyi Qin, Kuan Fang, Yuke Zhu, Li~Fei-Fei, and Silvio Savarese.
\newblock Keto: Learning keypoint representations for tool manipulation.
\newblock In \emph{2020 IEEE International Conference on Robotics and Automation (ICRA)}, pages 7278--7285. IEEE, 2020.

\bibitem[Rashid et~al.(2023)Rashid, Sharma, Kim, Kerr, Chen, Kanazawa, and Goldberg]{rashid2023language}
Adam Rashid, Satvik Sharma, Chung~Min Kim, Justin Kerr, Lawrence~Yunliang Chen, Angjoo Kanazawa, and Ken Goldberg.
\newblock Language embedded radiance fields for zero-shot task-oriented grasping.
\newblock In \emph{Conference on Robot Learning}, pages 178--200. PMLR, 2023.

\bibitem[Reed et~al.(2022)Reed, Zolna, Parisotto, Colmenarejo, Novikov, Barth-Maron, Gimenez, Sulsky, Kay, Springenberg, et~al.]{reed2022generalist}
Scott Reed, Konrad Zolna, Emilio Parisotto, Sergio~Gomez Colmenarejo, Alexander Novikov, Gabriel Barth-Maron, Mai Gimenez, Yury Sulsky, Jackie Kay, Jost~Tobias Springenberg, et~al.
\newblock A generalist agent.
\newblock \emph{arXiv preprint arXiv:2205.06175}, 2022.

\bibitem[Schmeckpeper et~al.(2019)Schmeckpeper, Xie, Rybkin, Tian, Daniilidis, Levine, and Finn]{schmeckpeper2019learning}
Karl Schmeckpeper, Annie Xie, Oleh Rybkin, Stephen Tian, Kostas Daniilidis, Sergey Levine, and Chelsea Finn.
\newblock Learning predictive models from observation and interaction, 2019.

\bibitem[Schmidt and Jiang(2024)]{schmidt2024learning}
Dominik Schmidt and Minqi Jiang.
\newblock Learning to act without actions.
\newblock In \emph{The Twelfth International Conference on Learning Representations}, 2024.
\newblock URL \url{https://openreview.net/forum?id=rvUq3cxpDF}.

\bibitem[Seita et~al.(2023)Seita, Wang, Shetty, Li, Erickson, and Held]{seita2023toolflownet}
Daniel Seita, Yufei Wang, Sarthak~J Shetty, Edward~Yao Li, Zackory Erickson, and David Held.
\newblock Toolflownet: Robotic manipulation with tools via predicting tool flow from point clouds.
\newblock In \emph{Conference on Robot Learning}, pages 1038--1049. PMLR, 2023.

\bibitem[Seo et~al.(2022)Seo, Lee, James, and Abbeel]{seo2022reinforcement}
Younggyo Seo, Kimin Lee, Stephen James, and Pieter Abbeel.
\newblock Reinforcement learning with action-free pre-training from videos, 2022.

\bibitem[Sermanet et~al.(2018{\natexlab{a}})Sermanet, Lynch, Chebotar, Hsu, Jang, Schaal, Levine, and Brain]{sermanet2017tcn}
Pierre Sermanet, Corey Lynch, Yevgen Chebotar, Jasmine Hsu, Eric Jang, Stefan Schaal, Sergey Levine, and Google Brain.
\newblock Time-contrastive networks: Self-supervised learning from video.
\newblock In \emph{2018 IEEE international conference on robotics and automation (ICRA)}, pages 1134--1141. IEEE, 2018{\natexlab{a}}.

\bibitem[Sermanet et~al.(2018{\natexlab{b}})Sermanet, Lynch, Chebotar, Hsu, Jang, Schaal, Levine, and Brain]{sermanet2018tcn}
Pierre Sermanet, Corey Lynch, Yevgen Chebotar, Jasmine Hsu, Eric Jang, Stefan Schaal, Sergey Levine, and Google Brain.
\newblock Time-contrastive networks: Self-supervised learning from video.
\newblock In \emph{2018 IEEE international conference on robotics and automation (ICRA)}, pages 1134--1141. IEEE, 2018{\natexlab{b}}.

\bibitem[Shao et~al.(2021)Shao, Migimatsu, Zhang, Yang, and Bohg]{shao2021concept2robot}
Lin Shao, Toki Migimatsu, Qiang Zhang, Karen Yang, and Jeannette Bohg.
\newblock Concept2robot: Learning manipulation concepts from instructions and human demonstrations.
\newblock \emph{The International Journal of Robotics Research}, 40\penalty0 (12-14):\penalty0 1419--1434, 2021.

\bibitem[Shaw et~al.(2023)Shaw, Bahl, and Pathak]{shaw2023videodex}
Kenneth Shaw, Shikhar Bahl, and Deepak Pathak.
\newblock Videodex: Learning dexterity from internet videos.
\newblock In \emph{Conference on Robot Learning}, pages 654--665. PMLR, 2023.

\bibitem[Torabi et~al.(2018)Torabi, Warnell, and Stone]{torabi2018behavioral}
Faraz Torabi, Garrett Warnell, and Peter Stone.
\newblock Behavioral cloning from observation, 2018.

\bibitem[Vecerik et~al.(2023)Vecerik, Doersch, Yang, Davchev, Aytar, Zhou, Hadsell, Agapito, and Scholz]{vecerik2023robotap}
Mel Vecerik, Carl Doersch, Yi~Yang, Todor Davchev, Yusuf Aytar, Guangyao Zhou, Raia Hadsell, Lourdes Agapito, and Jon Scholz.
\newblock Robotap: Tracking arbitrary points for few-shot visual imitation.
\newblock \emph{arXiv}, 2023.

\bibitem[Walke et~al.(2023)Walke, Black, Zhao, Vuong, Zheng, Hansen-Estruch, He, Myers, Kim, Du, et~al.]{walke2023bridgedata}
Homer~Rich Walke, Kevin Black, Tony~Z Zhao, Quan Vuong, Chongyi Zheng, Philippe Hansen-Estruch, Andre~Wang He, Vivek Myers, Moo~Jin Kim, Max Du, et~al.
\newblock Bridgedata v2: A dataset for robot learning at scale.
\newblock In \emph{Conference on Robot Learning}, pages 1723--1736. PMLR, 2023.

\bibitem[Wang et~al.(2023{\natexlab{a}})Wang, Fan, Sun, Zhang, Fei-Fei, Xu, Zhu, and Anandkumar]{wang2023mimicplay}
Chen Wang, Linxi Fan, Jiankai Sun, Ruohan Zhang, Li~Fei-Fei, Danfei Xu, Yuke Zhu, and Anima Anandkumar.
\newblock Mimicplay: Long-horizon imitation learning by watching human play.
\newblock In \emph{7th Annual Conference on Robot Learning}, 2023{\natexlab{a}}.

\bibitem[Wang et~al.(2023{\natexlab{b}})Wang, Chang, Cai, Li, Hariharan, Holynski, and Snavely]{wang2023tracking}
Qianqian Wang, Yen-Yu Chang, Ruojin Cai, Zhengqi Li, Bharath Hariharan, Aleksander Holynski, and Noah Snavely.
\newblock Tracking everything everywhere all at once, 2023{\natexlab{b}}.

\bibitem[Wen et~al.(2022)Wen, Qian, Lin, Teng, Jayaraman, and Gao]{wen2022primenet}
Chuan Wen, Jianing Qian, Jierui Lin, Jiaye Teng, Dinesh Jayaraman, and Yang Gao.
\newblock Fighting fire with fire: Avoiding dnn shortcuts through priming.
\newblock In \emph{International Conference on Machine Learning}, pages 23723--23750. PMLR, 2022.

\bibitem[Wu et~al.(2023)Wu, Shentu, Yi, Lin, and Abbeel]{wu2023gello}
Philipp Wu, Yide Shentu, Zhongke Yi, Xingyu Lin, and Pieter Abbeel.
\newblock Gello: A general, low-cost, and intuitive teleoperation framework for robot manipulators.
\newblock \emph{arXiv preprint arXiv:2309.13037}, 2023.

\bibitem[Xiong et~al.(2021)Xiong, Li, Chen, Bharadhwaj, Sinha, and Garg]{xiong2021learning}
Haoyu Xiong, Quanzhou Li, Yun-Chun Chen, Homanga Bharadhwaj, Samarth Sinha, and Animesh Garg.
\newblock Learning by watching: Physical imitation of manipulation skills from human videos.
\newblock In \emph{2021 IEEE/RSJ International Conference on Intelligent Robots and Systems (IROS)}, pages 7827--7834. IEEE, 2021.

\bibitem[Yang et~al.(2023)Yang, Du, Ghasemipour, Tompson, Schuurmans, and Abbeel]{yang2023learning}
Mengjiao Yang, Yilun Du, Kamyar Ghasemipour, Jonathan Tompson, Dale Schuurmans, and Pieter Abbeel.
\newblock Learning interactive real-world simulators.
\newblock \emph{arXiv preprint arXiv:2310.06114}, 2023.

\bibitem[Zhang et~al.(2018)Zhang, McCarthy, Jow, Lee, Chen, Goldberg, and Abbeel]{zhang2018deep}
Tianhao Zhang, Zoe McCarthy, Owen Jow, Dennis Lee, Xi~Chen, Ken Goldberg, and Pieter Abbeel.
\newblock Deep imitation learning for complex manipulation tasks from virtual reality teleoperation.
\newblock In \emph{2018 IEEE International Conference on Robotics and Automation (ICRA)}, pages 5628--5635. IEEE, 2018.

\bibitem[Zheng et~al.(2023)Zheng, Harley, Shen, Wetzstein, and Guibas]{zheng2023pointodyssey}
Yang Zheng, Adam~W Harley, Bokui Shen, Gordon Wetzstein, and Leonidas~J Guibas.
\newblock Pointodyssey: A large-scale synthetic dataset for long-term point tracking.
\newblock In \emph{Proceedings of the IEEE/CVF International Conference on Computer Vision}, pages 19855--19865, 2023.

\end{thebibliography}
